\documentclass[10pt,twocolumn,letterpaper]{article}

\usepackage[pagenumbers]{cvpr}     % To produce the REVIEW version
\usepackage{amsmath,amssymb}
\usepackage{algorithm}
\usepackage{algpseudocode}
\usepackage{graphicx}
\usepackage{placeins}
\usepackage{subcaption}
\usepackage{booktabs}
\usepackage{multirow}
\usepackage{indentfirst}
\usepackage{adjustbox}
\definecolor{cvprblue}{rgb}{0.21,0.49,0.74}
\usepackage[pagebackref,breaklinks,colorlinks,allcolors=cvprblue]{hyperref}

\makeatletter
\@ifpackageloaded{array}{}{\RequirePackage{array}}
\@ifpackageloaded{makecell}{}{\RequirePackage{makecell}}
\@ifpackageloaded{multirow}{}{\RequirePackage{multirow}}
\@ifpackageloaded{xcolor}{}{\RequirePackage{xcolor}}
\@ifpackageloaded{colortbl}{}{\RequirePackage{colortbl}}
\@ifpackageloaded{booktabs}{}{\RequirePackage{booktabs}}
\@ifpackageloaded{caption}{}{\RequirePackage{caption}}
\@ifpackageloaded{fix-cm}{}{\RequirePackage{fix-cm}}

\@ifundefined{TOneWidth}{\newlength{\TOneWidth}}{}
\@ifundefined{TOneMetricWidth}{\newlength{\TOneMetricWidth}}{}
\makeatother

\definecolor{TOneHeaderBase}{cmyk}{0.62,0.57,0.23,0}
\colorlet{TOneHeader}{TOneHeaderBase!20}
\colorlet{TOneStripe}{gray!10}
\definecolor{TOneRed}{cmyk}{0,0.77,0.87,0}
\definecolor{TOneDownColor}{cmyk}{0.85,0,0.33,0}

\providecommand{\TOneUp}[1]{}
\providecommand{\TOneDown}[1]{}
\renewcommand{\TOneUp}[1]{%
  $_{\color{TOneRed}\uparrow #1}$}
\renewcommand{\TOneDown}[1]{%
  $_{\color{TOneDownColor}\downarrow #1}$}

\DeclareMathSizes{5}{5}{3.5}{2.5}
\DeclareMathSizes{5.5}{5.5}{3.85}{2.75}
\DeclareMathSizes{6}{6}{4.2}{3}
\DeclareMathSizes{6.5}{6.5}{4.55}{3.25}

\def\confName{CVPR}
\def\confYear{2027}

\title{Psychometrics-Inspired Coreset Selection for Image Quality Assessment}
\author{%
  Susu Jiang\textsuperscript{1},
  Xuelin Shen\textsuperscript{2},
  and Baoliang CHEN\textsuperscript{3}\\[2pt]
  \textsuperscript{1}Shenzhen University\quad
  \textsuperscript{2}GUANGMING Laboratory\quad
  \textsuperscript{3}City University of Hong Kong
}

\begin{document}
\raggedbottom

\maketitle

\begin{abstract}
Collecting large-scale subjective annotations remains a critical bottleneck in training accurate image quality assessment (IQA) models. Coreset selection reduces this burden by selecting representative images before annotation. However, diversity in a visual feature space does not necessarily reflect diversity in perceptual quality, raising a fundamental question about which quality-related factors the representation should encode. We propose GRAM, a quality-aware representation learning framework inspired by the graded response model (GRM) from psychometrics. GRM describes how an individual's latent ability determines the probabilities of reaching successive assessment grades, with response sensitivity and grade thresholds governing this relationship.  In this paper, we find that this principle naturally aligns with perceptual quality assessment, where the impact of image degradation depends on image content and distortion characteristics. Building on this connection, GRAM models quality judgments through three complementary factors: latent quality, content-dependent response sensitivity, and distortion-dependent grade thresholds. These factors jointly determine the probabilities of attaining different quality grades and the resulting expected quality score.  The learned parameters provide a structured quality-aware representation that complements general visual features for coreset selection. Experiments on nine datasets demonstrate improvements across different selection strategies, downstream IQA models, and annotation budgets. The code will be made publicly available.

\end{abstract}

\section{Introduction}
\label{sec:introduction}

\begin{figure}[t]
    \centering
    \includegraphics[width=\linewidth]{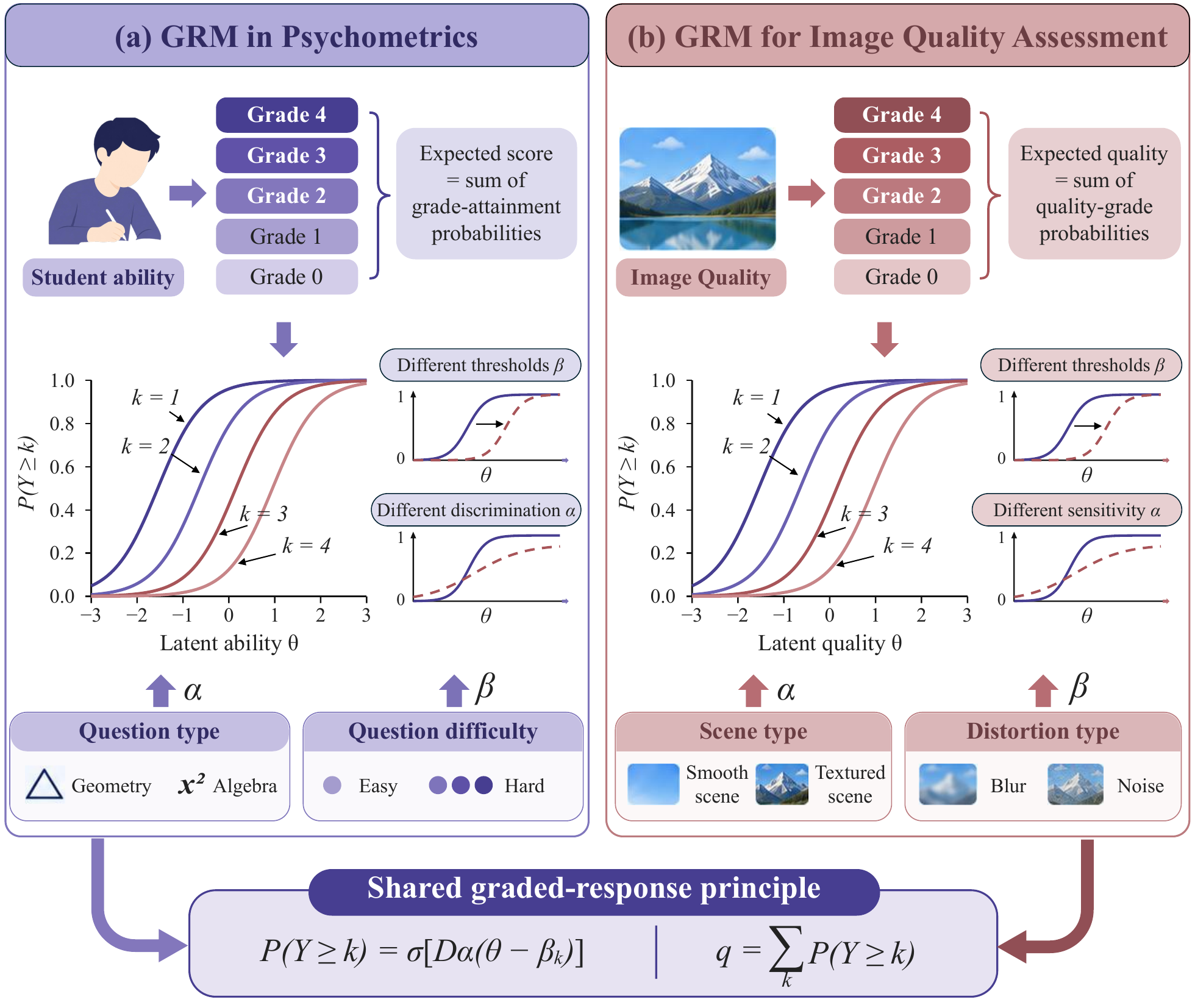}

\caption{
Shared graded-response principle for psychometrics and image quality assessment.
The graded response model (GRM) links a latent trait to ordinal judgments through
$P(Y \geq k) = \sigma[D\alpha(\theta - \beta_k)]$,
the probability of attaining at least grade $k$.
Here, $\sigma$ is the logistic sigmoid and $D$ is a fixed scaling factor.
\textbf{(a) Psychometrics.}
For a student with latent ability $\theta$, question difficulty is reflected in
the thresholds $\beta_k$: harder questions require greater ability to attain
the same grade and therefore shift the response curves to the right.
Question type, such as geometry or algebra, can affect discrimination $\alpha$,
which controls how sharply grade-attainment probabilities change with ability.
\textbf{(b) Image quality assessment.}
The same formulation treats $\theta$ as latent perceptual quality and $Y$
as an ordinal quality grade.
Scene content (e.g., smooth versus textured regions) and distortion type
(e.g., blur versus noise) can affect response sensitivity $\alpha$ and grade
thresholds $\beta_k$, reflecting differences in how readily degradations
are perceived and assigned quality grades.
In both settings, summing the cumulative probabilities gives the expected grade,
$q = \mathbb{E}[Y] = \sum_{k=1}^{K} P(Y \geq k)$,
converting graded responses into an expected score or quality prediction.
}
\label{fig:grm_comparison}
\label{fig:grm_motivation}

\end{figure}

An image quality assessment (IQA) model must learn to distinguish changes in perceptual quality across diverse contents and distortion conditions. With a limited annotation budget, the choice of training images determines which of these variations the model can observe. Coreset selection turns this choice into a subset selection problem: identify a small collection of images that preserves useful learning signals from a larger candidate pool. 

The effectiveness of coreset selection depends on how its representation defines sample relationships. Sophisticated selection strategies remain limited when the underlying representation overlooks differences relevant to the target task. For IQA, different distorted versions of the same content may remain close in a general visual feature space, even though their quality differences provide useful supervision. Moreover, images with similar quality scores may exhibit different perceptual characteristics due to variations in content and distortion. A representative subset thus needs to account for image content, distortion type, and distortion severity. Existing IQA coreset methods recognize the importance of perceptual information: PGCS partitions a perceptual representation space~\cite{nema2025holistic}, while Q-Diverse fuses content and quality distances for diversity-based sampling~\cite{nema2026diversity}. These advances motivate a further question: \textit{how can we learn a structured representation that captures the relationships underlying perceptual quality without subjective quality supervision?}

We draw inspiration from the graded response model (GRM) in psychometrics~\cite{samejima1969graded}. As shown in Fig.~\ref{fig:grm_motivation}, GRM describes how an underlying ability translates into graded assessment outcomes. Its central idea is that an observed score depends not only on the ability being assessed, but also on the characteristics of the assessment itself. For example, learners with the same mathematical ability may receive different grades on different questions as the questions vary in difficulty and sensitivity to ability. GRM accounts for these differences through three factors: latent ability, response discrimination, and grade thresholds. Together, they determine the probabilities of attaining successive grades and the resulting expected score. This formulation separates the underlying ability from the characteristics governing its graded evaluation.

Image quality assessment shares a similar structure. The perceptual impact of image degradation depends not only on distortion severity, but also on image content and distortion characteristics. For example, the same degree of blur may affect images with fine structures and smooth regions differently, while different distortion types may lead to different quality judgments. This motivates modeling image quality as a conditional graded response rather than directly mapping visual features to a single quality score. Following GRM, we introduce \textit{latent quality} to represent the underlying quality level of an image, \textit{response discrimination} to characterize content-dependent sensitivity, and \textit{grade thresholds} to describe distortion-dependent quality boundaries. These factors jointly determine the probabilities of attaining different quality grades and the resulting expected quality score. Importantly, images with similar expected quality scores can have different parameter configurations, providing complementary information for selecting representative training samples.

Based on this formulation, we propose \textbf{GRAM}, a \textbf{G}raded \textbf{R}esponse model-\textbf{A}ug\textbf{M}ented representation learning framework for IQA coreset selection. GRAM predicts image-conditioned GRM parameters and learns them from synthetically distorted images without subjective quality annotations. Severity-based quality ranking guides the learning of latent quality, while content and distortion relation constraints encourage response discrimination and grade thresholds to capture complementary information. Rather than relying on the predicted quality score, we adopt the learned parameters as a multidimensional quality-aware representation and fuse them with general visual features. The resulting representation serves as a plug-and-play input to existing coreset selection algorithms. Selected images are subsequently annotated and used to train  downstream IQA models.

We evaluate GRAM on nine datasets spanning synthetic distortions, authentic distortions, and AI-generated images, using four selection strategies and three downstream IQA models. Relative to selection with general features alone, GRAM improves downstream performance consistently, with larger average gains under limited annotation budgets. Parameter and subset analyses examine the learned relationships and show that changes in content, quality, and distortion coverage depend on the selection strategy.

% Our main contributions are as follows:
% \begin{itemize}
%     \item We propose GRAM, a psychometrics-inspired framework that connects latent distortion severity to ordered quality responses through content-related discrimination and distortion-related thresholds.
    
%     \item We develop a parameter learning method that uses severity-based ranking and content--distortion relation constraints to learn a compact selection representation without subjective quality supervision.
    
%     \item We construct a compact selection space that combines multidimensional GRM parameter representations with visual features, allowing existing coreset strategies to exploit differences in parameter configurations even among images with similar expected quality scores.
% \end{itemize}

\section{Related Work}
\label{sec:related_work}

\begin{figure*}[t]
    \centering
    \includegraphics[width=\textwidth]{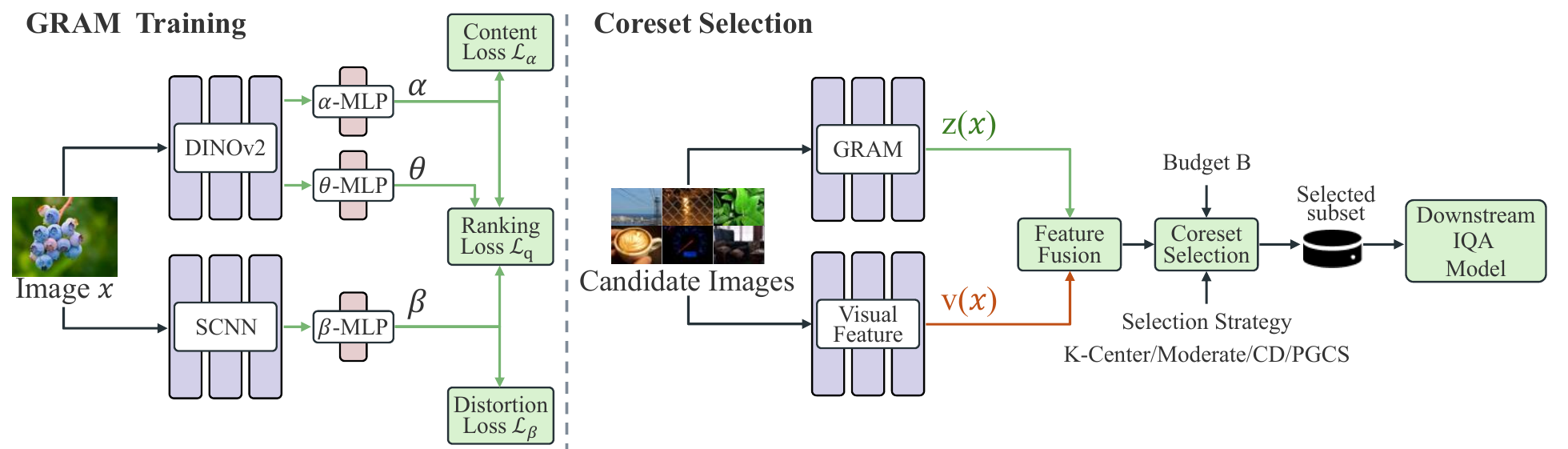}
    \caption{Overview of GRAM for quality-aware coreset selection. \textbf{Left: Training.} Frozen DINOv2 and SCNN encoders provide semantic and distortion features to predict latent quality $\theta$, discrimination $\alpha$, and ordered thresholds $\boldsymbol{\beta}$. Quality ranking jointly supervises all three branches through the predicted quality score, while content and distortion relation losses constrain $\alpha$ and $\boldsymbol{\beta}$, respectively. Training uses synthetically distorted images without subjective quality annotations. \textbf{Right: Coreset selection.} The GRM representation $\mathbf{z}(x)$ is fused with general visual features $\mathbf{v}(x)$ for different selection strategies under annotation budget $B$. Selected images are then annotated and used to train downstream IQA models.}
    \label{fig:grm_parameters}
\end{figure*}

\subsection{Coreset Selection}
\label{sec:related_coreset}

Coreset selection seeks representative subsets that reduce annotation and training costs while preserving downstream performance. K-Center~\cite{sener2018coreset} emphasizes feature-space coverage, and Moderate~\cite{xia2023moderate} selects samples with distance-based selection scores near the median. Contextual Diversity~\cite{agarwal2020contextual} assesses sample diversity using contextual information from model predictions. Other approaches construct subsets through gradient or distribution matching~\cite{killamsetty2021gradmatch,cui2026fast}. In short, these methods illustrate different criteria for selecting representative samples, including coverage, diversity, and gradient or distribution matching. Applying these methods to IQA requires accounting for perceptual quality. Perceptually Guided Coreset Selection (PGCS)~\cite{nema2025holistic} combines perceptual-space partitioning with distance-based sampling within each partition. Q-Diverse~\cite{nema2026diversity} fuses distances derived from content-aware and quality-aware features and selects diverse samples in the resulting spectral embedding. For AI-generated IQA, WCPC~\cite{shinde2026wcpc} uses Wasserstein Characteristic Divergence to select subsets that preserve dataset-level statistical characteristics. Building on these perceptually informed selection methods, we investigate whether explicitly learning relationships among quality, content, and distortion can provide complementary features for different selection strategies.

\subsection{Perceptual Representation Learning}
\label{sec:related_representation}

Pretrained visual representations provide a basis for measuring sample similarity. DINOv2~\cite{oquab2024dinov2} learns transferable visual features through large-scale self-supervised learning. IQA-oriented methods also seek to capture distortion and quality information through task-specific training objectives. DBCNN [29] uses bilinear pooling to fuse features pretrained for distortion type and severity classification with those pretrained for image classification. RankIQA~\cite{liu2017rankiqa} uses quality rankings of synthetically distorted images to learn features for quality prediction. ARNIQA~\cite{agnolucci2024arniqa} models a distortion manifold through contrastive learning, while Re-IQA~\cite{saha2023reiqa} learns separate content and quality representations. LIQE~\cite{zhang2023liqe} uses vision--language correspondence to jointly learn representations for quality assessment, scene classification, and distortion identification. These studies suggest that effective IQA coreset selection benefits from representations that capture the factors underlying perceptual quality.

\section{Method}
\label{sec:method}

% \begin{figure*}[t]
%     \centering
%     \includegraphics[width=\textwidth]{Figure/frame.pdf}
%     \caption{Overview of GRAM for quality-aware coreset selection. \textbf{Left: Training.} Frozen DINOv2 and SCNN encoders provide semantic and distortion features to predict latent quality $\theta$, discrimination $\alpha$, and ordered thresholds $\boldsymbol{\beta}$. Quality ranking jointly supervises all three branches through the predicted quality score, while content and distortion relation losses constrain $\alpha$ and $\boldsymbol{\beta}$, respectively. Training uses synthetically distorted images without subjective quality annotations. \textbf{Right: Coreset selection.} The GRM representation $\mathbf{z}(x)$ is fused with general visual features $\mathbf{v}(x)$ for different selection strategies under annotation budget $B$. Selected images are then annotated and used to train downstream IQA models.}
%     \label{fig:grm_parameters}
% \end{figure*}

\subsection{Preliminaries}
\label{sec:grm}

The graded response model (GRM)~\cite{samejima1969graded} is a classical model of ordered assessment responses based on learner ability and question characteristics. For example, learners with the same ability may receive different grades on questions of different difficulty.

Consider a question with response grade $Y\in\{0,\ldots,K\}$. Its response is described by learner ability $\theta$, question discrimination $\alpha>0$, and ordered grade thresholds $\boldsymbol{\beta}=(\beta_1,\ldots,\beta_K)$ with $\beta_1<\cdots<\beta_K$. The conditional expected score is
\begin{equation}
Q(\theta,\alpha,\boldsymbol{\beta})
= \mathbb{E}[Y\mid\theta,\alpha,\boldsymbol{\beta}]
= \sum_{r=0}^{K} r\,\pi_r,
% 保留原来的 \label{...}
\end{equation}
where $\pi_r=P(Y=r\mid\theta,\alpha,\boldsymbol{\beta})$
is the probability of receiving grade $r$.
Since a response of grade $r$ reaches the first $r$ levels,
defining $p_k=P(Y\geq k\mid\theta,\alpha,\boldsymbol{\beta})$
gives
\begin{equation}
Q(\theta,\alpha,\boldsymbol{\beta})=\sum_{k=1}^{K}p_k.
\label{eq:grm_score_cumulative}
\end{equation}
Score prediction thus reduces to modeling the cumulative probabilities of attaining successive grades. In GRM, the probabilities are modeled as
\begin{equation}
p_k=\sigma\!\left[D\alpha(\theta-\beta_k)\right],
\quad k=1,\ldots,K.
\label{eq:grm_cumulative}
\end{equation}
Here, $\sigma(t)=(1+e^{-t})^{-1}$ and $D$ is a fixed scaling constant~\cite{camilli1994scaling}. Discrimination $\alpha$ controls how sensitively the response varies with ability, while the thresholds specify the ability requirements for successive grades. Their question-specific values account for differences in response sensitivity and grade requirements across questions.

This structure, in turn, provides a principled framework for modeling ordered image quality judgments across diverse scene and distortion conditions. For an image \(x\), we interpret \(\theta(x)\) as latent quality, representing its underlying quality level. Higher latent quality corresponds to a greater likelihood of attaining higher quality grades under fixed response parameters. We associate \(\alpha(x)\) with content-dependent response sensitivity and \(\boldsymbol{\beta}(x)\) with distortion-dependent quality boundaries. Using these image-conditioned parameters in the same response model, we obtain the expected quality score \(q(x)=Q(\theta(x),\alpha(x),\boldsymbol{\beta}(x))\). Specifically, the same latent quality  can yield different quality scores under different sensitivities and thresholds, while different parameter configurations can produce similar scores.

\subsection{Learning Quality-aware Representations}
\label{sec:grm_representation}

To learn the image-to-parameter mapping without the need for
subjective quality annotations, we exploit the inherent quality
ordering in synthetically distorted datasets.
Fig.~\ref{fig:grm_parameters} outlines the parameter learning framework.

\noindent\textbf{Parameter prediction.}
We predict the GRM parameters using complementary semantic
and distortion features: \mbox{DINOv2~\cite{oquab2024dinov2}}
features predict $\theta(x)$ and $\alpha(x)$, while features
from SCNN~\cite{zhang2020dbcnn}, pretrained to classify synthetic distortion
types and severity levels, predict the thresholds.
Both encoders remain frozen.

Three prediction branches map these features to the GRM parameters, with $K=4$ (five quality grades) by default.
Two separate MLPs use L2-normalized \mbox{DINOv2} CLS features to predict $\theta(x)$ and $\alpha(x)$, with scaled sigmoid
outputs in $[1,5]$ and $[1,3]$, respectively.
Global average pooling of the SCNN feature feeds a threshold
branch that predicts a base threshold $b(x)$ and positive
increments $\Delta_2(x),\ldots,\Delta_4(x)$.
The thresholds are constructed as
\begin{equation}
\beta_k(x)=b(x)+\sum_{t=2}^{k}\Delta_t(x),
\quad k=1,\ldots,4.
\label{eq:ordered_threshold_prediction}
\end{equation}
For $k=1$, the sum is empty and $\beta_1(x)=b(x)$.
This parameterization guarantees ordered thresholds.
Only the prediction branches are updated during training.

\noindent\textbf{Training relations.}
Let $x_{c,d,s}$ denote a synthetic image with reference content~$c$, distortion type~$d$, and severity level $s\in\{1,\ldots,5\}$. Larger~$s$ indicates stronger distortion. Each training block contains two contents, two distortion types, and all five severity levels, yielding 20 images. These blocks provide quality orders within each content--distortion pair and matched-severity comparisons for learning content and distortion relations.

\noindent\textbf{Quality ranking.}
For fixed content and distortion type, increasing synthetic severity should decrease quality. We impose this ordering on the conditional quality scores $q_s=q(x_{c,d,s})$ by minimizing
\begin{equation}
\mathcal{L}_q
=\mathbb{E}_{c,d}\!\left[
\frac{\sum_{1\leq s<t\leq5}
 w_{st}\log\!\left(1+e^{-(q_s-q_t)/\tau}\right)}
 {\sum_{1\leq s<t\leq5}w_{st}}
\right],
\label{eq:quality_ranking}
\end{equation}
where $w_{st}=(t-s)/4$ gives greater weight to larger severity gaps, $\tau>0$ is a temperature, and the expectation averages content--distortion pairs. This loss jointly updates all three parameter branches through $q(x)$.

\noindent\textbf{Content and distortion relations.}
Quality ranking constrains the combined response of the parameters. We additionally use relational supervision to encourage discrimination to retain content-related information and thresholds to learn the distortion-related information. For anchor $x_a=x_{c,d,s}$, define $x_{\mathrm{C}}=x_{c,d',s}$ with the same content and a different distortion, and $x_{\mathrm{D}}=x_{c',d,s}$ with the same distortion and a different content, where $c'\ne c$ and $d'\ne d$. All three images have matched severity.

For discrimination, $x_{\mathrm{C}}$ is the positive and $x_{\mathrm{D}}$ is the negative. The content relation loss is
\begin{equation}
\mathcal{L}_{\alpha}
=\mathbb{E}_{\mathcal{T}}\!\bigl\{
 [|\alpha_a-\alpha_{\mathrm{C}}|
 -|\alpha_a-\alpha_{\mathrm{D}}|+m_{\alpha}]_+
\bigr\}.
\label{eq:grm_content_triplet}
\end{equation}
For thresholds, we reverse these roles and use
\begin{equation}
\mathcal{L}_{\beta}
=\mathbb{E}_{\mathcal{T}}\!\bigl\{
 [d_{\beta}(\boldsymbol{\beta}_a,\boldsymbol{\beta}_{\mathrm{D}})
 -d_{\beta}(\boldsymbol{\beta}_a,\boldsymbol{\beta}_{\mathrm{C}})
 +m_{\beta}]_+
\bigr\},
\label{eq:grm_distortion_triplet}
\end{equation}
where $\mathcal{T}$ contains the matched-severity triplets in each block, $[u]_+=\max(u,0)$, $m_{\alpha},m_{\beta}>0$ are margins, and $d_{\beta}$ is the root-mean-square distance between threshold vectors. Subscripts identify predictions for the corresponding images.

\noindent\textbf{Training objective and representation.}
The complete objective is
\begin{equation}
\mathcal{L}=\mathcal{L}_q+\mathcal{L}_{\alpha}+\mathcal{L}_{\beta}.
\label{eq:overall_loss}
\end{equation}
The ranking objective couples the parameters through quality prediction, while the relational objectives encourage complementary content and distortion information. After training, we freeze the model and retain the full parameter configuration,
\begin{equation}
\mathbf{z}(x)=
[\theta(x),\alpha(x),\beta_1(x),\ldots,\beta_4(x)]^{\mathsf T}
\in\mathbb{R}^{6}.
\label{eq:grm_representation}
\end{equation}
This representation enables selection to account for learned parameter differences among images with similar expected quality scores.

\subsection{Representation Fusion and Coreset Selection}
\label{sec:subset_selection}

We combine the learned parameter representation with general visual features to measure both semantic similarity and quality-related sample differences. Given an unannotated candidate set $\mathcal{D}_{\mathrm{train}}=\{x_i\}_{i=1}^{N}$ and selection ratio $0<\rho\leq1$, our objective is to select $B=\max(1,\lfloor\rho N\rfloor)$ images for subsequent annotation and IQA training.

\noindent\textbf{Representation fusion.}
For each image $x$, we extract an L2-normalized visual feature
$\mathbf{v}(x)\in\mathbb{R}^{d_v}$ from a pretrained backbone
in the VGG~\cite{simonyan2015vgg}, ResNet~\cite{he2016resnet}, or DINO families.
DINOv2~\cite{oquab2024dinov2} serves as the default encoder
($d_v=768$) for its strong semantic representations and transferability.
We use the six-dimensional GRM parameter representation
$\mathbf{z}(x)$ defined in Eq.~\eqref{eq:grm_representation}
as complementary features.

To balance the contributions of the two branches to distance
computations, we construct three coordinates from each branch.
The visual feature is reduced through a PCA mapping $f_v$
and then standardized coordinate-wise to obtain
$\widetilde{\mathbf{v}}(x)\in\mathbb{R}^{3}$.
For the GRM branch, the threshold vector $\boldsymbol{\beta}(x)$
is first standardized coordinate-wise and then compressed into
a single coordinate through a UMAP mapping
$f_{\beta}$~\cite{mcinnes2018umap}.
The resulting scalar is standardized again to obtain
$\widetilde{\beta}(x)$.
Latent quality and discrimination are separately standardized
to obtain $\widetilde{\theta}(x)$ and $\widetilde{\alpha}(x)$,
yielding three GRM coordinates. This processing gives both branches equal dimensionality
and places their coordinates on comparable scales.
We concatenate them into the fused selection representation:
\begin{equation}
\mathbf{h}(x)=
\bigl[
\widetilde{\mathbf{v}}(x)^{\mathsf T},
\widetilde{\theta}(x),
\widetilde{\alpha}(x),
\widetilde{\beta}(x)
\bigr]^{\mathsf T}
\in\mathbb{R}^{6}.
\label{eq:fused_representation}
\end{equation}

\noindent\textbf{Coreset selection and downstream training.}
The resulting features serve as a plug-and-play representation
for different coreset selection strategies, enabling them to
exploit quality-relevant sample relationships.
As summarized in Algorithm~\ref{alg:coreset_selection},
given features $H=\{\mathbf{h}(x_i)\}_{i=1}^{N}$ and annotation
budget $B$, a strategy $\mathcal{A}$ returns indices
$I=\mathcal{A}(H,B)$ and the coreset
$\mathcal{S}=\{x_i:i\in I\}$ without accessing candidate MOS labels.
The selected images are then annotated and used to train
downstream IQA models to improve their predictive performance
under limited annotation budgets.

\begin{algorithm}[t]
\caption{Coreset Selection and Downstream IQA Training}
\label{alg:coreset_selection}
\begin{algorithmic}[1]
\Require Candidate images
$\mathcal{D}_{\mathrm{train}}=\{x_i\}_{i=1}^{N}$,
fused features $H$, selection ratio $\rho$, selection strategy $\mathcal{A}$, IQA model $\mathcal{M}$
\Ensure Coreset $\mathcal{S}$ and trained IQA model $\mathcal{M}^{*}$
\State $B \gets \max(1,\lfloor \rho N \rfloor)$
\State $I \gets \mathcal{A}(H,B)$
\State $\mathcal{S} \gets \{x_i:i\in I\}$
\State Obtain subjective quality scores $y_i$ for $i\in I$
\State $\mathcal{D}_{\mathcal{S}}
       \gets \{(x_i,y_i):i\in I\}$
\State $\mathcal{M}^{*}
       \gets \operatorname{Train}(\mathcal{M},
                                 \mathcal{D}_{\mathcal{S}})$
\State \Return $\mathcal{S},\mathcal{M}^{*}$
\end{algorithmic}
\end{algorithm}

\section{Experiments}
\label{sec:experiment}

% ==================== TABLE 1 - double column, independent ====================
\begin{table*}[!t]
\captionsetup{font=small,aboveskip=0pt,belowskip=2pt,width=\TOneWidth,justification=justified,singlelinecheck=false}
\centering
\caption{DBCNN results: median test SRCC/PLCC over ten splits. Mean: nine-dataset average. Base: DINOv2-768D; Ours: \mbox{DINOv2+GRM.} Bold: Ours $>$ Base; subscripts: $100(\mathrm{Ours}-\mathrm{Base})/\mathrm{Base}$ (\%; $\uparrow$ gain, $\downarrow$ drop). KADID-10K\textsuperscript{*}: separately formed GRM training and downstream test splits may share reference images or distorted samples; upstream ranking uses synthetic distortion severity.}
\label{tab:main_dbcnn}
\begingroup
\fontsize{5}{6.8}\selectfont
\setlength{\tabcolsep}{1.9pt}
\renewcommand{\arraystretch}{1}
\setlength{\arrayrulewidth}{0.25pt}
\definecolor{TOneHeaderBase}{cmyk}{0.62,0.57,0.23,0}
\colorlet{TOneHeader}{TOneHeaderBase!20}
\colorlet{TOneStripe}{gray!10}
\definecolor{TOneRed}{cmyk}{0,0.77,0.87,0}
\definecolor{TOneDownColor}{cmyk}{0.85,0,0.33,0}
\def\TOneUp#1{$_{\color{TOneRed}\uparrow #1}$}
\def\TOneDown#1{$_{\color{TOneDownColor}\downarrow #1}$}
\setlength{\TOneMetricWidth}{\dimexpr(\TOneWidth-63pt-26\tabcolsep-\arrayrulewidth)/10\relax}
\begin{tabular}{W{c}{19pt}W{c}{20pt}W{l}{24pt}|*{10}{W{c}{\TOneMetricWidth}}}
\Xhline{0.7pt}
\rowcolor{TOneHeader}
 & & &
\multicolumn{2}{c}{\textbf{CLIVE}} &
\multicolumn{2}{c}{\textbf{KonIQ-10K}} &
\multicolumn{2}{c}{\textbf{AIGCIQA2023}} &
\multicolumn{2}{c}{\textbf{AGIQA-3K}} &
\multicolumn{2}{c}{\textbf{PKU-AIGIQA-4K}} \\

\rowcolor{TOneHeader}
\multirow{-2}{*}{\textbf{Feature}} &
\multirow{-2}{*}{\textbf{Budget}} &
\multicolumn{1}{c|}{\multirow{-2}{*}{\textbf{Method}}} &
\textbf{SRCC} & \textbf{PLCC} &
\textbf{SRCC} & \textbf{PLCC} &
\textbf{SRCC} & \textbf{PLCC} &
\textbf{SRCC} & \textbf{PLCC} &
\textbf{SRCC} & \textbf{PLCC} \\
\Xhline{0.7pt}

% ==================== Base: 5% ====================
 &  & K-Center & $0.5211$ & $0.5458$ & $0.6228$ & $0.6497$ & $0.6774$ & $0.6750$ & $0.6169$ & $0.6458$ & $0.4630$ & $0.4848$ \\
\rowcolor{TOneStripe}
 &  & CD & $0.5036$ & $0.5593$ & $0.6366$ & $0.6843$ & $0.6993$ & $0.7071$ & $0.6983$ & $0.7578$ & $0.4625$ & $0.4658$ \\
 &  & Moderate & $0.5185$ & $0.6015$ & $0.6321$ & $0.6516$ & $0.6398$ & $0.6445$ & $0.6518$ & $0.7130$ & $0.4841$ & $0.4853$ \\
\rowcolor{TOneStripe}
 & \multirow{-4}{*}{5\%} & PGCS & $0.4590$ & $0.4812$ & $0.6125$ & $0.6459$ & $0.6509$ & $0.6641$ & $0.6506$ & $0.7152$ & $0.4899$ & $0.4758$ \\
\cline{2-13}

% ==================== Base: 10% ====================
 &  & K-Center & $0.6623$ & $0.6827$ & $0.7022$ & $0.7311$ & $0.7484$ & $0.7632$ & $0.7142$ & $0.7636$ & $0.5343$ & $0.5382$ \\
\rowcolor{TOneStripe}
 &  & CD & $0.6327$ & $0.6752$ & $0.6876$ & $0.7268$ & $0.7428$ & $0.7554$ & $0.7625$ & $0.8163$ & $0.5253$ & $0.5288$ \\
 &  & Moderate & $0.6214$ & $0.6593$ & $0.7034$ & $0.7306$ & $0.6967$ & $0.7123$ & $0.7134$ & $0.7850$ & $0.5276$ & $0.5263$ \\
\rowcolor{TOneStripe}
 & \multirow{-4}{*}{10\%} & PGCS & $0.6062$ & $0.6292$ & $0.6950$ & $0.7203$ & $0.7224$ & $0.7338$ & $0.7158$ & $0.7785$ & $0.5229$ & $0.5265$ \\
\cline{2-13}

% ==================== Base: 20% ====================
 &  & K-Center & $0.6988$ & $0.7293$ & $0.7463$ & $0.7770$ & $0.7789$ & $0.7938$ & $0.7787$ & $0.8292$ & $0.5706$ & $0.5831$ \\
\rowcolor{TOneStripe}
 &  & CD & $0.6936$ & $0.7447$ & $0.7326$ & $0.7728$ & $0.7721$ & $0.7866$ & $0.7921$ & $0.8491$ & $0.5866$ & $0.5913$ \\
 &  & Moderate & $0.6726$ & $0.7174$ & $0.7482$ & $0.7772$ & $0.7507$ & $0.7654$ & $0.7462$ & $0.8045$ & $0.5555$ & $0.5632$ \\
\rowcolor{TOneStripe}
 & \multirow{-4}{*}{20\%} & PGCS & $0.6626$ & $0.7113$ & $0.7445$ & $0.7737$ & $0.7652$ & $0.7765$ & $0.7467$ & $0.8070$ & $0.5590$ & $0.5668$ \\
\cline{2-13}

% ==================== Base: 30% ====================
 &  & K-Center & $0.7283$ & $0.7632$ & $0.7660$ & $0.7981$ & $0.7879$ & $0.8081$ & $0.7965$ & $0.8514$ & $0.6077$ & $0.6158$ \\
\rowcolor{TOneStripe}
 &  & CD & $0.7072$ & $0.7682$ & $0.7603$ & $0.7936$ & $0.7940$ & $0.8088$ & $0.7984$ & $0.8596$ & $0.6206$ & $0.6276$ \\
 &  & Moderate & $0.7069$ & $0.7509$ & $0.7649$ & $0.7929$ & $0.7769$ & $0.7924$ & $0.7722$ & $0.8290$ & $0.5951$ & $0.5967$ \\
\rowcolor{TOneStripe}
 & \multirow{-4}{*}{30\%} & PGCS & $0.7142$ & $0.7512$ & $0.7655$ & $0.7930$ & $0.7860$ & $0.7979$ & $0.7677$ & $0.8284$ & $0.6014$ & $0.5992$ \\
\cline{2-13}

% ==================== Base: 50% ====================
 &  & K-Center & $0.7620$ & $0.8005$ & $0.7846$ & $0.8164$ & $0.8033$ & $0.8223$ & $0.8141$ & $0.8676$ & $0.6498$ & $0.6614$ \\
\rowcolor{TOneStripe}
 &  & CD & $0.7549$ & $0.8010$ & $0.7852$ & $0.8160$ & $0.8055$ & $0.8253$ & $0.8149$ & $0.8745$ & $0.6592$ & $0.6654$ \\
 &  & Moderate & $0.7347$ & $0.7891$ & $0.7847$ & $0.8136$ & $0.7904$ & $0.8103$ & $0.7904$ & $0.8504$ & $0.6373$ & $0.6387$ \\
\rowcolor{TOneStripe}
\multirow{-20}{*}{Base} & \multirow{-4}{*}{50\%} & PGCS & $0.7618$ & $0.8018$ & $0.7850$ & $0.8142$ & $0.8014$ & $0.8148$ & $0.7916$ & $0.8485$ & $0.6355$ & $0.6318$ \\
\hline

% ==================== Ours: 5% ====================
 &  & K-Center & $\mathbf{0.5817}$\TOneUp{11.63} & $\mathbf{0.6304}$\TOneUp{15.50} & $0.6188$\TOneDown{0.64} & $\mathbf{0.6765}$\TOneUp{4.12} & $\mathbf{0.7091}$\TOneUp{4.68} & $\mathbf{0.7158}$\TOneUp{6.04} & $\mathbf{0.7319}$\TOneUp{18.64} & $\mathbf{0.7717}$\TOneUp{19.50} & $\mathbf{0.4631}$\TOneUp{0.02} & $0.4634$\TOneDown{4.41} \\
\rowcolor{TOneStripe}
 &  & CD & $\mathbf{0.5876}$\TOneUp{16.68} & $\mathbf{0.6050}$\TOneUp{8.17} & $0.6294$\TOneDown{1.13} & $0.6808$\TOneDown{0.51} & $0.6983$\TOneDown{0.14} & $\mathbf{0.7128}$\TOneUp{0.81} & $\mathbf{0.7220}$\TOneUp{3.39} & $\mathbf{0.7684}$\TOneUp{1.40} & $0.4472$\TOneDown{3.31} & $0.4469$\TOneDown{4.06} \\
 &  & Moderate & $\mathbf{0.5534}$\TOneUp{6.73} & $0.5947$\TOneDown{1.13} & $\mathbf{0.6350}$\TOneUp{0.46} & $\mathbf{0.6740}$\TOneUp{3.44} & $\mathbf{0.7099}$\TOneUp{10.96} & $\mathbf{0.7245}$\TOneUp{12.41} & $\mathbf{0.6869}$\TOneUp{5.39} & $\mathbf{0.7291}$\TOneUp{2.26} & $\mathbf{0.4866}$\TOneUp{0.52} & $\mathbf{0.4870}$\TOneUp{0.35} \\
\rowcolor{TOneStripe}
 & \multirow{-4}{*}{5\%} & PGCS & $\mathbf{0.5038}$\TOneUp{9.76} & $\mathbf{0.5564}$\TOneUp{15.63} & $\mathbf{0.6368}$\TOneUp{3.97} & $\mathbf{0.6728}$\TOneUp{4.16} & $\mathbf{0.6884}$\TOneUp{5.76} & $\mathbf{0.6958}$\TOneUp{4.77} & $\mathbf{0.6812}$\TOneUp{4.70} & $\mathbf{0.7280}$\TOneUp{1.79} & $0.4789$\TOneDown{2.25} & $\mathbf{0.4897}$\TOneUp{2.92} \\
\cline{2-13}

% ==================== Ours: 10% ====================
 &  & K-Center & $0.6478$\TOneDown{2.19} & $0.6802$\TOneDown{0.37} & $0.6838$\TOneDown{2.62} & $0.7298$\TOneDown{0.18} & $\mathbf{0.7543}$\TOneUp{0.79} & $\mathbf{0.7709}$\TOneUp{1.01} & $\mathbf{0.7642}$\TOneUp{7.00} & $\mathbf{0.8154}$\TOneUp{6.78} & $0.5293$\TOneDown{0.94} & $0.5359$\TOneDown{0.43} \\
\rowcolor{TOneStripe}
 &  & CD & $\mathbf{0.6670}$\TOneUp{5.42} & $\mathbf{0.7143}$\TOneUp{5.79} & $0.6746$\TOneDown{1.89} & $0.7232$\TOneDown{0.50} & $\mathbf{0.7539}$\TOneUp{1.49} & $\mathbf{0.7697}$\TOneUp{1.89} & $\mathbf{0.7651}$\TOneUp{0.34} & $0.8100$\TOneDown{0.77} & $\mathbf{0.5378}$\TOneUp{2.38} & $\mathbf{0.5331}$\TOneUp{0.81} \\
 &  & Moderate & $\mathbf{0.6488}$\TOneUp{4.41} & $\mathbf{0.6908}$\TOneUp{4.78} & $\mathbf{0.7038}$\TOneUp{0.06} & $\mathbf{0.7403}$\TOneUp{1.33} & $\mathbf{0.7380}$\TOneUp{5.93} & $\mathbf{0.7530}$\TOneUp{5.71} & $\mathbf{0.7429}$\TOneUp{4.14} & $\mathbf{0.7955}$\TOneUp{1.34} & $\mathbf{0.5295}$\TOneUp{0.36} & $\mathbf{0.5383}$\TOneUp{2.28} \\
\rowcolor{TOneStripe}
 & \multirow{-4}{*}{10\%} & PGCS & $\mathbf{0.6281}$\TOneUp{3.61} & $\mathbf{0.6729}$\TOneUp{6.95} & $\mathbf{0.7012}$\TOneUp{0.89} & $\mathbf{0.7334}$\TOneUp{1.82} & $\mathbf{0.7365}$\TOneUp{1.95} & $\mathbf{0.7522}$\TOneUp{2.51} & $\mathbf{0.7442}$\TOneUp{3.97} & $\mathbf{0.7959}$\TOneUp{2.24} & $\mathbf{0.5314}$\TOneUp{1.63} & $\mathbf{0.5444}$\TOneUp{3.40} \\
\cline{2-13}

% ==================== Ours: 20% ====================
 &  & K-Center & $\mathbf{0.7160}$\TOneUp{2.46} & $\mathbf{0.7594}$\TOneUp{4.13} & $0.7308$\TOneDown{2.08} & $0.7719$\TOneDown{0.66} & $\mathbf{0.7841}$\TOneUp{0.67} & $\mathbf{0.8036}$\TOneUp{1.23} & $\mathbf{0.7960}$\TOneUp{2.22} & $\mathbf{0.8489}$\TOneUp{2.38} & $\mathbf{0.5933}$\TOneUp{3.98} & $\mathbf{0.5990}$\TOneUp{2.73} \\
\rowcolor{TOneStripe}
 &  & CD & $\mathbf{0.7009}$\TOneUp{1.05} & $\mathbf{0.7626}$\TOneUp{2.40} & $0.7297$\TOneDown{0.40} & $\mathbf{0.7746}$\TOneUp{0.23} & $\mathbf{0.7868}$\TOneUp{1.90} & $\mathbf{0.8033}$\TOneUp{2.12} & $\mathbf{0.7943}$\TOneUp{0.28} & $\mathbf{0.8506}$\TOneUp{0.18} & $0.5859$\TOneDown{0.12} & $0.5824$\TOneDown{1.51} \\
 &  & Moderate & $\mathbf{0.7008}$\TOneUp{4.19} & $\mathbf{0.7560}$\TOneUp{5.38} & $\mathbf{0.7495}$\TOneUp{0.17} & $\mathbf{0.7807}$\TOneUp{0.45} & $\mathbf{0.7735}$\TOneUp{3.04} & $\mathbf{0.7840}$\TOneUp{2.43} & $\mathbf{0.7847}$\TOneUp{5.16} & $\mathbf{0.8354}$\TOneUp{3.84} & $\mathbf{0.5782}$\TOneUp{4.09} & $\mathbf{0.5871}$\TOneUp{4.24} \\
\rowcolor{TOneStripe}
 & \multirow{-4}{*}{20\%} & PGCS & $\mathbf{0.6941}$\TOneUp{4.75} & $\mathbf{0.7589}$\TOneUp{6.69} & $\mathbf{0.7498}$\TOneUp{0.71} & $\mathbf{0.7815}$\TOneUp{1.01} & $\mathbf{0.7684}$\TOneUp{0.42} & $\mathbf{0.7833}$\TOneUp{0.88} & $\mathbf{0.7776}$\TOneUp{4.14} & $\mathbf{0.8340}$\TOneUp{3.35} & $\mathbf{0.5796}$\TOneUp{3.69} & $\mathbf{0.5860}$\TOneUp{3.39} \\
\cline{2-13}

% ==================== Ours: 30% ====================
 &  & K-Center & $\mathbf{0.7359}$\TOneUp{1.04} & $\mathbf{0.7854}$\TOneUp{2.91} & $0.7574$\TOneDown{1.12} & $0.7958$\TOneDown{0.29} & $\mathbf{0.7960}$\TOneUp{1.03} & $\mathbf{0.8113}$\TOneUp{0.40} & $\mathbf{0.8049}$\TOneUp{1.05} & $\mathbf{0.8630}$\TOneUp{1.36} & $\mathbf{0.6304}$\TOneUp{3.74} & $\mathbf{0.6294}$\TOneUp{2.21} \\
\rowcolor{TOneStripe}
 &  & CD & $\mathbf{0.7359}$\TOneUp{4.06} & $\mathbf{0.7921}$\TOneUp{3.11} & $0.7570$\TOneDown{0.43} & $\mathbf{0.7963}$\TOneUp{0.34} & $\mathbf{0.7951}$\TOneUp{0.14} & $0.8067$\TOneDown{0.26} & $\mathbf{0.8052}$\TOneUp{0.85} & $0.8565$\TOneDown{0.36} & $\mathbf{0.6217}$\TOneUp{0.18} & $0.6158$\TOneDown{1.88} \\
 &  & Moderate & $\mathbf{0.7331}$\TOneUp{3.71} & $\mathbf{0.7855}$\TOneUp{4.61} & $\mathbf{0.7714}$\TOneUp{0.85} & $\mathbf{0.7998}$\TOneUp{0.87} & $\mathbf{0.7918}$\TOneUp{1.92} & $\mathbf{0.8062}$\TOneUp{1.74} & $\mathbf{0.7960}$\TOneUp{3.08} & $\mathbf{0.8501}$\TOneUp{2.55} & $\mathbf{0.6211}$\TOneUp{4.37} & $\mathbf{0.6248}$\TOneUp{4.71} \\
\rowcolor{TOneStripe}
 & \multirow{-4}{*}{30\%} & PGCS & $\mathbf{0.7254}$\TOneUp{1.57} & $\mathbf{0.7780}$\TOneUp{3.57} & $\mathbf{0.7707}$\TOneUp{0.68} & $\mathbf{0.7998}$\TOneUp{0.86} & $\mathbf{0.7862}$\TOneUp{0.03} & $\mathbf{0.8043}$\TOneUp{0.80} & $\mathbf{0.7980}$\TOneUp{3.95} & $\mathbf{0.8511}$\TOneUp{2.74} & $\mathbf{0.6166}$\TOneUp{2.53} & $\mathbf{0.6184}$\TOneUp{3.20} \\
\cline{2-13}

% ==================== Ours: 50% ====================
 &  & K-Center & $\mathbf{0.7854}$\TOneUp{3.07} & $\mathbf{0.8159}$\TOneUp{1.92} & $0.7831$\TOneDown{0.19} & $0.8155$\TOneDown{0.11} & $\mathbf{0.8079}$\TOneUp{0.57} & $0.8203$\TOneDown{0.24} & $\mathbf{0.8147}$\TOneUp{0.07} & $\mathbf{0.8737}$\TOneUp{0.70} & $\mathbf{0.6656}$\TOneUp{2.43} & $\mathbf{0.6686}$\TOneUp{1.09} \\
\rowcolor{TOneStripe}
 &  & CD & $\mathbf{0.7815}$\TOneUp{3.52} & $\mathbf{0.8174}$\TOneUp{2.05} & $0.7825$\TOneDown{0.34} & $\mathbf{0.8166}$\TOneUp{0.07} & $\mathbf{0.8096}$\TOneUp{0.51} & $0.8250$\TOneDown{0.04} & $\mathbf{0.8190}$\TOneUp{0.50} & $0.8736$\TOneDown{0.10} & $\mathbf{0.6804}$\TOneUp{3.22} & $\mathbf{0.6748}$\TOneUp{1.41} \\
 &  & Moderate & $\mathbf{0.7690}$\TOneUp{4.67} & $\mathbf{0.8161}$\TOneUp{3.42} & $\mathbf{0.7882}$\TOneUp{0.45} & $\mathbf{0.8167}$\TOneUp{0.38} & $\mathbf{0.8024}$\TOneUp{1.52} & $\mathbf{0.8129}$\TOneUp{0.32} & $\mathbf{0.8121}$\TOneUp{2.75} & $\mathbf{0.8655}$\TOneUp{1.78} & $\mathbf{0.6610}$\TOneUp{3.72} & $\mathbf{0.6626}$\TOneUp{3.74} \\
\rowcolor{TOneStripe}
\multirow{-20}{*}{Ours} & \multirow{-4}{*}{50\%} & PGCS & $\mathbf{0.7744}$\TOneUp{1.65} & $\mathbf{0.8175}$\TOneUp{1.96} & $\mathbf{0.7889}$\TOneUp{0.50} & $\mathbf{0.8166}$\TOneUp{0.29} & $0.7983$\TOneDown{0.39} & $0.8125$\TOneDown{0.28} & $\mathbf{0.8182}$\TOneUp{3.36} & $\mathbf{0.8694}$\TOneUp{2.46} & $\mathbf{0.6568}$\TOneUp{3.35} & $\mathbf{0.6643}$\TOneUp{5.14} \\
\hline

% ==================== Full: no vertical rule ====================
\multicolumn{1}{c}{} & \rule[-2pt]{0pt}{9pt}Full & \multicolumn{1}{l}{} & $0.7987$ & $0.8343$ & $0.8074$ & $0.8360$ & $0.8206$ & $0.8374$ & $0.8265$ & $0.8831$ & $0.7125$ & $0.7184$ \\
\Xhline{0.7pt}
\end{tabular}\par
\endgroup
\par\vspace{4pt}
\centering
\begingroup
\fontsize{5}{6.8}\selectfont
\setlength{\tabcolsep}{1.9pt}
\renewcommand{\arraystretch}{1}
\setlength{\arrayrulewidth}{0.25pt}
\definecolor{TOneHeaderBase}{cmyk}{0.62,0.57,0.23,0}
\colorlet{TOneHeader}{TOneHeaderBase!20}
\colorlet{TOneStripe}{gray!10}
\definecolor{TOneRed}{cmyk}{0,0.77,0.87,0}
\definecolor{TOneDownColor}{cmyk}{0.85,0,0.33,0}
\def\TOneUp#1{$_{\color{TOneRed}\uparrow #1}$}
\def\TOneDown#1{$_{\color{TOneDownColor}\downarrow #1}$}
\setlength{\TOneMetricWidth}{\dimexpr(\TOneWidth-63pt-26\tabcolsep-\arrayrulewidth)/10\relax}
\begin{tabular}{W{c}{19pt}W{c}{20pt}W{l}{24pt}|*{10}{W{c}{\TOneMetricWidth}}}
\Xhline{0.7pt}
\rowcolor{TOneHeader}
 & & & \multicolumn{2}{c}{\textbf{LIVE}} & \multicolumn{2}{c}{\textbf{CSIQ}} & \multicolumn{2}{c}{\textbf{TID2013}} & \multicolumn{2}{c}{\textbf{KADID-10K}\textsuperscript{*}} & \multicolumn{2}{c}{\textbf{Mean}} \\
\rowcolor{TOneHeader}
\multirow{-2}{*}{\textbf{Feature}} & \multirow{-2}{*}{\textbf{Budget}} & \multicolumn{1}{c|}{\multirow{-2}{*}{\textbf{Method}}} & \textbf{SRCC} & \textbf{PLCC} & \textbf{SRCC} & \textbf{PLCC} & \textbf{SRCC} & \textbf{PLCC} & \textbf{SRCC} & \textbf{PLCC} & \textbf{SRCC} & \textbf{PLCC} \\
\Xhline{0.7pt}

% ==================== Base: 5% ====================
 &  & K-Center & $0.8191$ & $0.8425$ & $0.7583$ & $0.8353$ & $0.5869$ & $0.6859$ & $0.5861$ & $0.6063$ & $0.6280$ & $0.6635$ \\
\rowcolor{TOneStripe}
 &  & CD & $0.8383$ & $0.8525$ & $0.7992$ & $0.8350$ & $0.6842$ & $0.7577$ & $0.6166$ & $0.6334$ & $0.6598$ & $0.6948$ \\
 &  & Moderate & $0.8345$ & $0.8334$ & $0.8136$ & $0.8367$ & $0.6569$ & $0.6765$ & $0.5778$ & $0.5936$ & $0.6455$ & $0.6707$ \\
\rowcolor{TOneStripe}
 & \multirow{-4}{*}{5\%} & PGCS & $0.8312$ & $0.8347$ & $0.8102$ & $0.8557$ & $0.7100$ & $0.7351$ & $0.5967$ & $0.6158$ & $0.6457$ & $0.6693$ \\
\cline{2-13}

% ==================== Base: 10% ====================
 &  & K-Center & $0.8508$ & $0.8688$ & $0.7781$ & $0.8605$ & $0.6347$ & $0.7158$ & $0.6438$ & $0.6594$ & $0.6965$ & $0.7315$ \\
\rowcolor{TOneStripe}
 &  & CD & $0.8892$ & $0.8806$ & $0.8304$ & $0.8841$ & $0.7268$ & $0.8026$ & $0.6549$ & $0.6726$ & $0.7169$ & $0.7492$ \\
 &  & Moderate & $0.8905$ & $0.8933$ & $0.8459$ & $0.8779$ & $0.6973$ & $0.7162$ & $0.6121$ & $0.6234$ & $0.7009$ & $0.7249$ \\
\rowcolor{TOneStripe}
 & \multirow{-4}{*}{10\%} & PGCS & $0.9017$ & $0.9037$ & $0.8389$ & $0.8825$ & $0.7595$ & $0.7891$ & $0.6334$ & $0.6467$ & $0.7106$ & $0.7345$ \\
\cline{2-13}

% ==================== Base: 20% ====================
 &  & K-Center & $0.8909$ & $0.9045$ & $0.8616$ & $0.9028$ & $0.7480$ & $0.8085$ & $0.7006$ & $0.7134$ & $0.7527$ & $0.7824$ \\
\rowcolor{TOneStripe}
 &  & CD & $0.9186$ & $0.9238$ & $0.8737$ & $0.9095$ & $0.7952$ & $0.8505$ & $0.7143$ & $0.7260$ & $0.7643$ & $0.7949$ \\
 &  & Moderate & $0.9185$ & $0.9266$ & $0.8715$ & $0.9011$ & $0.7409$ & $0.7652$ & $0.6579$ & $0.6697$ & $0.7402$ & $0.7656$ \\
\rowcolor{TOneStripe}
 & \multirow{-4}{*}{20\%} & PGCS & $0.9243$ & $0.9310$ & $0.8944$ & $0.9152$ & $0.8005$ & $0.8295$ & $0.6870$ & $0.6952$ & $0.7538$ & $0.7785$ \\
\cline{2-13}

% ==================== Base: 30% ====================
 &  & K-Center & $0.9172$ & $0.9295$ & $0.8921$ & $0.9291$ & $0.8298$ & $0.8726$ & $0.7171$ & $0.7271$ & $0.7825$ & $0.8105$ \\
\rowcolor{TOneStripe}
 &  & CD & $0.9365$ & $0.9437$ & $0.8978$ & $0.9308$ & $0.8544$ & $0.8936$ & $0.7170$ & $0.7322$ & $0.7874$ & $0.8176$ \\
 &  & Moderate & $0.9295$ & $0.9267$ & $0.8903$ & $0.9119$ & $0.7565$ & $0.7784$ & $0.6846$ & $0.6977$ & $0.7641$ & $0.7863$ \\
\rowcolor{TOneStripe}
 & \multirow{-4}{*}{30\%} & PGCS & $0.9332$ & $0.9397$ & $0.9242$ & $0.9432$ & $0.8335$ & $0.8610$ & $0.7154$ & $0.7228$ & $0.7823$ & $0.8040$ \\
\cline{2-13}

% ==================== Base: 50% ====================
 &  & K-Center & $0.9490$ & $0.9581$ & $0.9492$ & $0.9651$ & $0.9078$ & $0.9258$ & $0.7623$ & $0.7656$ & $0.8202$ & $0.8425$ \\
\rowcolor{TOneStripe}
 &  & CD & $0.9562$ & $0.9599$ & $0.9428$ & $0.9639$ & $0.9156$ & $0.9319$ & $0.7601$ & $0.7649$ & $0.8216$ & $0.8448$ \\
 &  & Moderate & $0.9431$ & $0.9442$ & $0.9242$ & $0.9385$ & $0.8365$ & $0.8630$ & $0.7368$ & $0.7421$ & $0.7976$ & $0.8211$ \\
\rowcolor{TOneStripe}
\multirow{-20}{*}{Base} & \multirow{-4}{*}{50\%} & PGCS & $0.9411$ & $0.9458$ & $0.9596$ & $0.9697$ & $0.8904$ & $0.9005$ & $0.7510$ & $0.7548$ & $0.8130$ & $0.8313$ \\
\hline

% ==================== Ours: 5% ====================
 &  & K-Center & $\mathbf{0.8306}$\TOneUp{1.40} & $0.8306$\TOneDown{1.41} & $\mathbf{0.8086}$\TOneUp{6.63} & $\mathbf{0.8634}$\TOneUp{3.36} & $\mathbf{0.7368}$\TOneUp{25.54} & $\mathbf{0.7856}$\TOneUp{14.54} & $\mathbf{0.6482}$\TOneUp{10.60} & $\mathbf{0.6586}$\TOneUp{8.63} & $\mathbf{0.6810}$\TOneUp{8.44} & $\mathbf{0.7107}$\TOneUp{7.12} \\
\rowcolor{TOneStripe}
 &  & CD & $\mathbf{0.8536}$\TOneUp{1.83} & $0.8417$\TOneDown{1.27} & $\mathbf{0.8256}$\TOneUp{3.30} & $\mathbf{0.8671}$\TOneUp{3.84} & $\mathbf{0.7443}$\TOneUp{8.78} & $\mathbf{0.7920}$\TOneUp{4.53} & $\mathbf{0.6454}$\TOneUp{4.67} & $\mathbf{0.6593}$\TOneUp{4.09} & $\mathbf{0.6837}$\TOneUp{3.62} & $\mathbf{0.7082}$\TOneUp{1.94} \\
 &  & Moderate & $0.7990$\TOneDown{4.25} & $0.8100$\TOneDown{2.81} & $0.7864$\TOneDown{3.34} & $\mathbf{0.8436}$\TOneUp{0.82} & $\mathbf{0.6867}$\TOneUp{4.54} & $\mathbf{0.7385}$\TOneUp{9.16} & $\mathbf{0.5892}$\TOneUp{1.97} & $\mathbf{0.6038}$\TOneUp{1.72} & $\mathbf{0.6592}$\TOneUp{2.13} & $\mathbf{0.6895}$\TOneUp{2.80} \\
\rowcolor{TOneStripe}
 & \multirow{-4}{*}{5\%} & PGCS & $0.8299$\TOneDown{0.16} & $0.8288$\TOneDown{0.71} & $0.8084$\TOneDown{0.22} & $0.8484$\TOneDown{0.85} & $0.7090$\TOneDown{0.14} & $\mathbf{0.7447}$\TOneUp{1.31} & $0.5865$\TOneDown{1.71} & $0.6086$\TOneDown{1.17} & $\mathbf{0.6581}$\TOneUp{1.93} & $\mathbf{0.6859}$\TOneUp{2.49} \\
\cline{2-13}

% ==================== Ours: 10% ====================
 &  & K-Center & $\mathbf{0.9096}$\TOneUp{6.91} & $\mathbf{0.9163}$\TOneUp{5.47} & $\mathbf{0.8658}$\TOneUp{11.27} & $\mathbf{0.8939}$\TOneUp{3.88} & $\mathbf{0.8137}$\TOneUp{28.20} & $\mathbf{0.8511}$\TOneUp{18.90} & $\mathbf{0.6922}$\TOneUp{7.52} & $\mathbf{0.7007}$\TOneUp{6.26} & $\mathbf{0.7401}$\TOneUp{6.25} & $\mathbf{0.7660}$\TOneUp{4.72} \\
\rowcolor{TOneStripe}
 &  & CD & $\mathbf{0.8978}$\TOneUp{0.97} & $\mathbf{0.9002}$\TOneUp{2.23} & $\mathbf{0.8813}$\TOneUp{6.13} & $\mathbf{0.9059}$\TOneUp{2.47} & $\mathbf{0.8305}$\TOneUp{14.27} & $\mathbf{0.8618}$\TOneUp{7.38} & $\mathbf{0.6876}$\TOneUp{4.99} & $\mathbf{0.6941}$\TOneUp{3.20} & $\mathbf{0.7440}$\TOneUp{3.77} & $\mathbf{0.7680}$\TOneUp{2.52} \\
 &  & Moderate & $0.8835$\TOneDown{0.79} & $0.8873$\TOneDown{0.67} & $\mathbf{0.8617}$\TOneUp{1.87} & $\mathbf{0.8887}$\TOneUp{1.23} & $\mathbf{0.7600}$\TOneUp{8.99} & $\mathbf{0.7872}$\TOneUp{9.91} & $\mathbf{0.6213}$\TOneUp{1.50} & $\mathbf{0.6391}$\TOneUp{2.52} & $\mathbf{0.7211}$\TOneUp{2.87} & $\mathbf{0.7467}$\TOneUp{3.00} \\
\rowcolor{TOneStripe}
 & \multirow{-4}{*}{10\%} & PGCS & $0.8867$\TOneDown{1.66} & $\mathbf{0.9045}$\TOneUp{0.09} & $\mathbf{0.8533}$\TOneUp{1.72} & $0.8811$\TOneDown{0.16} & $\mathbf{0.7697}$\TOneUp{1.34} & $\mathbf{0.8066}$\TOneUp{2.22} & $0.6310$\TOneDown{0.38} & $0.6459$\TOneDown{0.12} & $\mathbf{0.7202}$\TOneUp{1.35} & $\mathbf{0.7485}$\TOneUp{1.92} \\
\cline{2-13}

% ==================== Ours: 20% ====================
 &  & K-Center & $\mathbf{0.9419}$\TOneUp{5.72} & $\mathbf{0.9433}$\TOneUp{4.29} & $\mathbf{0.9222}$\TOneUp{7.03} & $\mathbf{0.9433}$\TOneUp{4.49} & $\mathbf{0.8798}$\TOneUp{17.62} & $\mathbf{0.9034}$\TOneUp{11.74} & $\mathbf{0.7312}$\TOneUp{4.37} & $\mathbf{0.7373}$\TOneUp{3.35} & $\mathbf{0.7884}$\TOneUp{4.74} & $\mathbf{0.8122}$\TOneUp{3.81} \\
\rowcolor{TOneStripe}
 &  & CD & $\mathbf{0.9452}$\TOneUp{2.90} & $\mathbf{0.9472}$\TOneUp{2.53} & $\mathbf{0.9175}$\TOneUp{5.01} & $\mathbf{0.9378}$\TOneUp{3.11} & $\mathbf{0.8813}$\TOneUp{10.83} & $\mathbf{0.9069}$\TOneUp{6.63} & $\mathbf{0.7315}$\TOneUp{2.41} & $\mathbf{0.7388}$\TOneUp{1.76} & $\mathbf{0.7859}$\TOneUp{2.82} & $\mathbf{0.8116}$\TOneUp{2.10} \\
 &  & Moderate & $0.9183$\TOneDown{0.02} & $0.9208$\TOneDown{0.63} & $\mathbf{0.8949}$\TOneUp{2.69} & $\mathbf{0.9146}$\TOneUp{1.50} & $\mathbf{0.7976}$\TOneUp{7.65} & $\mathbf{0.8349}$\TOneUp{9.11} & $\mathbf{0.6783}$\TOneUp{3.10} & $\mathbf{0.6899}$\TOneUp{3.02} & $\mathbf{0.7640}$\TOneUp{3.21} & $\mathbf{0.7893}$\TOneUp{3.09} \\
\rowcolor{TOneStripe}
 & \multirow{-4}{*}{20\%} & PGCS & $0.9239$\TOneDown{0.04} & $0.9288$\TOneDown{0.24} & $\mathbf{0.8973}$\TOneUp{0.32} & $\mathbf{0.9206}$\TOneUp{0.59} & $\mathbf{0.8018}$\TOneUp{0.16} & $0.8275$\TOneDown{0.24} & $0.6783$\TOneDown{1.27} & $0.6905$\TOneDown{0.68} & $\mathbf{0.7634}$\TOneUp{1.28} & $\mathbf{0.7901}$\TOneUp{1.50} \\
\cline{2-13}

% ==================== Ours: 30% ====================
 &  & K-Center & $\mathbf{0.9546}$\TOneUp{4.08} & $\mathbf{0.9583}$\TOneUp{3.10} & $\mathbf{0.9422}$\TOneUp{5.62} & $\mathbf{0.9594}$\TOneUp{3.26} & $\mathbf{0.9082}$\TOneUp{9.45} & $\mathbf{0.9239}$\TOneUp{5.88} & $\mathbf{0.7498}$\TOneUp{4.56} & $\mathbf{0.7543}$\TOneUp{3.74} & $\mathbf{0.8088}$\TOneUp{3.36} & $\mathbf{0.8312}$\TOneUp{2.55} \\
\rowcolor{TOneStripe}
 &  & CD & $\mathbf{0.9549}$\TOneUp{1.96} & $\mathbf{0.9581}$\TOneUp{1.53} & $\mathbf{0.9489}$\TOneUp{5.69} & $\mathbf{0.9608}$\TOneUp{3.22} & $\mathbf{0.8969}$\TOneUp{4.97} & $\mathbf{0.9187}$\TOneUp{2.81} & $\mathbf{0.7489}$\TOneUp{4.45} & $\mathbf{0.7546}$\TOneUp{3.06} & $\mathbf{0.8072}$\TOneUp{2.52} & $\mathbf{0.8288}$\TOneUp{1.38} \\
 &  & Moderate & $\mathbf{0.9359}$\TOneUp{0.69} & $\mathbf{0.9434}$\TOneUp{1.80} & $\mathbf{0.9178}$\TOneUp{3.09} & $\mathbf{0.9392}$\TOneUp{2.99} & $\mathbf{0.8261}$\TOneUp{9.20} & $\mathbf{0.8582}$\TOneUp{10.25} & $\mathbf{0.7098}$\TOneUp{3.68} & $\mathbf{0.7193}$\TOneUp{3.10} & $\mathbf{0.7892}$\TOneUp{3.29} & $\mathbf{0.8141}$\TOneUp{3.53} \\
\rowcolor{TOneStripe}
 & \multirow{-4}{*}{30\%} & PGCS & $\mathbf{0.9392}$\TOneUp{0.64} & $\mathbf{0.9438}$\TOneUp{0.44} & $0.9232$\TOneDown{0.11} & $\mathbf{0.9436}$\TOneUp{0.04} & $\mathbf{0.8398}$\TOneUp{0.76} & $0.8562$\TOneDown{0.56} & $\mathbf{0.7164}$\TOneUp{0.14} & $\mathbf{0.7234}$\TOneUp{0.08} & $\mathbf{0.7906}$\TOneUp{1.06} & $\mathbf{0.8132}$\TOneUp{1.14} \\
\cline{2-13}

% ==================== Ours: 50% ====================
 &  & K-Center & $\mathbf{0.9708}$\TOneUp{2.30} & $\mathbf{0.9718}$\TOneUp{1.43} & $\mathbf{0.9655}$\TOneUp{1.72} & $\mathbf{0.9746}$\TOneUp{0.98} & $\mathbf{0.9244}$\TOneUp{1.83} & $\mathbf{0.9386}$\TOneUp{1.38} & $\mathbf{0.7714}$\TOneUp{1.19} & $\mathbf{0.7736}$\TOneUp{1.04} & $\mathbf{0.8321}$\TOneUp{1.45} & $\mathbf{0.8503}$\TOneUp{0.92} \\
\rowcolor{TOneStripe}
 &  & CD & $\mathbf{0.9654}$\TOneUp{0.96} & $\mathbf{0.9688}$\TOneUp{0.93} & $\mathbf{0.9654}$\TOneUp{2.40} & $\mathbf{0.9750}$\TOneUp{1.15} & $\mathbf{0.9218}$\TOneUp{0.68} & $\mathbf{0.9337}$\TOneUp{0.19} & $\mathbf{0.7734}$\TOneUp{1.75} & $\mathbf{0.7739}$\TOneUp{1.18} & $\mathbf{0.8332}$\TOneUp{1.41} & $\mathbf{0.8510}$\TOneUp{0.74} \\
 &  & Moderate & $\mathbf{0.9574}$\TOneUp{1.52} & $\mathbf{0.9598}$\TOneUp{1.65} & $\mathbf{0.9497}$\TOneUp{2.76} & $\mathbf{0.9627}$\TOneUp{2.58} & $\mathbf{0.8877}$\TOneUp{6.12} & $\mathbf{0.9028}$\TOneUp{4.61} & $\mathbf{0.7661}$\TOneUp{3.98} & $\mathbf{0.7615}$\TOneUp{2.61} & $\mathbf{0.8215}$\TOneUp{3.00} & $\mathbf{0.8401}$\TOneUp{2.31} \\
\rowcolor{TOneStripe}
\multirow{-20}{*}{Ours} & \multirow{-4}{*}{50\%} & PGCS & $\mathbf{0.9600}$\TOneUp{2.01} & $\mathbf{0.9627}$\TOneUp{1.79} & $0.9525$\TOneDown{0.74} & $0.9643$\TOneDown{0.56} & $0.8838$\TOneDown{0.74} & $\mathbf{0.9060}$\TOneUp{0.61} & $\mathbf{0.7660}$\TOneUp{2.00} & $\mathbf{0.7617}$\TOneUp{0.91} & $\mathbf{0.8221}$\TOneUp{1.11} & $\mathbf{0.8417}$\TOneUp{1.24} \\
\hline

% ==================== Full: no vertical rule ====================
\multicolumn{1}{c}{} & \rule[-2pt]{0pt}{9pt}Full & \multicolumn{1}{l}{} & $0.9767$ & $0.9785$ & $0.9791$ & $0.9826$ & $0.9429$ & $0.9472$ & $0.8070$ & $0.7980$ & $0.8524$ & $0.8684$ \\
\Xhline{0.7pt}
\end{tabular}\par
\endgroup
\end{table*}
\subsection{Experimental Setup}
\label{sec:experimental_setup}

 We evaluate on synthetic-distortion datasets LIVE IQA~\cite{sheikh2006live}, CSIQ~\cite{larson2010csiq}, TID2013~\cite{ponomarenko2015tid2013}, and KADID-10K~\cite{lin2019kadid}; authentic-distortion datasets CLIVE~\cite{ghadiyaram2016clive} and KonIQ-10K~\cite{hosu2020koniq}; and generated-image datasets AGIQA-3K~\cite{li2024agiqa}, AIGCIQA2023~\cite{wang2024aigciqa}, and PKU-AIGIQA-4K~\cite{yuan2025pkuaigiqa}. Quality scores are mapped to $[1,5]$. We use image-level stratified splits of 70\% training, 10\% validation, and 20\% testing, with ten splits and identical partitions across comparisons. The selection baseline uses L2-normalized DINOv2-768D features. We evaluate K-Center~\cite{sener2018coreset}, PGCS~\cite{nema2025holistic}, Moderate~\cite{xia2023moderate}, and CD~\cite{agarwal2020contextual} at training budgets of 5\%, 10\%, 20\%, 30\%, and 50\%. Full uses all training images.  We train MANIQA~\cite{yang2022maniqa}, LIQE~\cite{zhang2023liqe}, and DBCNN~\cite{zhang2020dbcnn}. Results are the median test SRCC and PLCC over ten splits.
A fixed GRM trained on KADID-10K is used throughout. Upstream and downstream KADID-10K splits are constructed independently. Detailed data preparation, representation and selection settings, and training configurations are provided in the supplementary material.

\begin{table*}[!t]
\centering
\captionsetup{font=small,aboveskip=0pt,belowskip=3pt,
    width=\linewidth,justification=justified,singlelinecheck=false}
\caption{Results with K-Center selection at a 5\% budget across three downstream IQA models. Dataset grouping, representations, and notation follow Table~\ref{tab:main_dbcnn}. Mean equally averages the scores across all nine datasets. KADID-10K\textsuperscript{*}: see Table~\ref{tab:main_dbcnn}.}
\label{tab:kcenter_three_models}
\begingroup
\fontsize{6}{7.2}\selectfont
\setlength{\tabcolsep}{2pt}
\renewcommand{\arraystretch}{1.15}
\setlength{\arrayrulewidth}{0.25pt}
\definecolor{KTHeaderBase}{cmyk}{0.62,0.57,0.23,0}
\colorlet{KTHeader}{KTHeaderBase!20}
\colorlet{KTStripe}{gray!10}
\definecolor{KTUpColor}{cmyk}{0,0.77,0.87,0}
\definecolor{KTDownColor}{cmyk}{0.85,0,0.33,0}

% 12 columns: 24pt + 28pt + 10 metric columns + 24 padding units + 1 rule.
% Both panels therefore occupy exactly the current double-column text width.
\edef\KTMetricWidth{\the\dimexpr(\linewidth-52pt-24\tabcolsep-\arrayrulewidth)/10\relax}
% Include the complete score AND change annotation in the cell's width.
\newcommand{\KTCell}[1]{\adjustbox{max width=\dimexpr\KTMetricWidth-1pt\relax}{#1}}
\newcommand{\KTBase}[1]{\KTCell{$#1$}}
\newcommand{\KTGain}[2]{\KTCell{$\mathbf{#1}_{\color{KTUpColor}\uparrow #2}$}}
\newcommand{\KTLoss}[2]{\KTCell{$#1_{\color{KTDownColor}\downarrow #2}$}}

\begin{tabular}{W{c}{24pt}W{l}{28pt}|*{10}{W{c}{\KTMetricWidth}}}
\toprule
\rowcolor{KTHeader}
 & & \multicolumn{2}{c}{\textbf{CLIVE}} & \multicolumn{2}{c}{\textbf{KonIQ-10K}} & \multicolumn{2}{c}{\textbf{AIGCIQA2023}} & \multicolumn{2}{c}{\textbf{AGIQA-3K}} & \multicolumn{2}{c}{\textbf{PKU-AIGIQA-4K}} \\
\rowcolor{KTHeader}
\multirow{-2}{*}{\textbf{Feature}} & \multirow{-2}{*}{\textbf{Model}} &
\textbf{SRCC} & \textbf{PLCC} & \textbf{SRCC} & \textbf{PLCC} & \textbf{SRCC} & \textbf{PLCC} & \textbf{SRCC} & \textbf{PLCC} & \textbf{SRCC} & \textbf{PLCC} \\
\midrule
 & MANIQA
& \KTBase{0.5398} & \KTBase{0.5360} & \KTBase{0.8338} & \KTBase{0.8512} & \KTBase{0.6826}
& \KTBase{0.6869} & \KTBase{0.7250} & \KTBase{0.7680} & \KTBase{0.4809} & \KTBase{0.4782} \\
\rowcolor{KTStripe}
 & LIQE
& \KTBase{0.7477} & \KTBase{0.7236} & \KTBase{0.8529} & \KTBase{0.7880} & \KTBase{0.7644}
& \KTBase{0.7784} & \KTBase{0.7947} & \KTBase{0.8134} & \KTBase{0.5470} & \KTBase{0.5622} \\
\multirow{-3}{*}{Base} & DBCNN
& \KTBase{0.5211} & \KTBase{0.5458} & \KTBase{0.6228} & \KTBase{0.6497} & \KTBase{0.6774}
& \KTBase{0.6750} & \KTBase{0.6169} & \KTBase{0.6458} & \KTBase{0.4630} & \KTBase{0.4848} \\
\midrule
 & MANIQA
& \KTGain{0.6335}{17.36} & \KTGain{0.6585}{22.85} & \KTLoss{0.8303}{0.42} & \KTGain{0.8601}{1.05} & \KTGain{0.7125}{4.38}
& \KTGain{0.7058}{2.75} & \KTGain{0.7807}{7.68} & \KTGain{0.8407}{9.47} & \KTGain{0.4870}{1.27} & \KTGain{0.4875}{1.94} \\
\rowcolor{KTStripe}
 & LIQE
& \KTGain{0.7762}{3.81} & \KTGain{0.7599}{5.02} & \KTGain{0.8584}{0.64} & \KTGain{0.8593}{9.05} & \KTGain{0.7940}{3.87}
& \KTGain{0.8089}{3.92} & \KTGain{0.8272}{4.09} & \KTGain{0.8534}{4.92} & \KTGain{0.5635}{3.02} & \KTLoss{0.5514}{1.92} \\
\multirow{-3}{*}{Ours} & DBCNN
& \KTGain{0.5817}{11.63} & \KTGain{0.6304}{15.50} & \KTLoss{0.6188}{0.64} & \KTGain{0.6765}{4.12} & \KTGain{0.7091}{4.68}
& \KTGain{0.7158}{6.04} & \KTGain{0.7319}{18.64} & \KTGain{0.7717}{19.50} & \KTGain{0.4631}{0.02} & \KTLoss{0.4634}{4.41} \\
\bottomrule
\end{tabular}\par
\vspace{5pt}

\begin{tabular}{W{c}{24pt}W{l}{28pt}|*{10}{W{c}{\KTMetricWidth}}}
\toprule
\rowcolor{KTHeader}
 & & \multicolumn{2}{c}{\textbf{LIVE}} & \multicolumn{2}{c}{\textbf{CSIQ}} & \multicolumn{2}{c}{\textbf{TID2013}} & \multicolumn{2}{c}{\textbf{KADID-10K}\textsuperscript{*}} & \multicolumn{2}{c}{\textbf{Mean}} \\
\rowcolor{KTHeader}
\multirow{-2}{*}{\textbf{Feature}} & \multirow{-2}{*}{\textbf{Model}} &
\textbf{SRCC} & \textbf{PLCC} & \textbf{SRCC} & \textbf{PLCC} & \textbf{SRCC} & \textbf{PLCC} & \textbf{SRCC} & \textbf{PLCC} & \textbf{SRCC} & \textbf{PLCC} \\
\midrule
 & MANIQA
& \KTBase{0.7332} & \KTBase{0.7744} & \KTBase{0.5646} & \KTBase{0.6585} & \KTBase{0.5313}
& \KTBase{0.6200} & \KTBase{0.6438} & \KTBase{0.6475} & \KTBase{0.6372} & \KTBase{0.6690} \\
\rowcolor{KTStripe}
 & LIQE
& \KTBase{0.8837} & \KTBase{0.9010} & \KTBase{0.8003} & \KTBase{0.8509} & \KTBase{0.6499}
& \KTBase{0.6655} & \KTBase{0.6978} & \KTBase{0.6452} & \KTBase{0.7487} & \KTBase{0.7476} \\
\multirow{-3}{*}{Base} & DBCNN
& \KTBase{0.8191} & \KTBase{0.8425} & \KTBase{0.7583} & \KTBase{0.8353} & \KTBase{0.5869}
& \KTBase{0.6859} & \KTBase{0.5861} & \KTBase{0.6063} & \KTBase{0.6280} & \KTBase{0.6635} \\
\midrule
 & MANIQA
& \KTGain{0.8199}{11.82} & \KTGain{0.8027}{3.65} & \KTGain{0.7268}{28.73} & \KTGain{0.7601}{15.43} & \KTGain{0.6769}{27.40}
& \KTGain{0.7488}{20.77} & \KTGain{0.8470}{31.56} & \KTGain{0.8579}{32.49} & \KTGain{0.7238}{13.59} & \KTGain{0.7469}{11.65} \\
\rowcolor{KTStripe}
 & LIQE
& \KTGain{0.9044}{2.34} & \KTGain{0.9023}{0.14} & \KTGain{0.8387}{4.80} & \KTGain{0.8759}{2.94} & \KTGain{0.7672}{18.05}
& \KTGain{0.8050}{20.96} & \KTGain{0.8499}{21.80} & \KTGain{0.8461}{31.14} & \KTGain{0.7977}{6.55} & \KTGain{0.8069}{7.94} \\
\multirow{-3}{*}{Ours} & DBCNN
& \KTGain{0.8306}{1.40} & \KTLoss{0.8306}{1.41} & \KTGain{0.8086}{6.63} & \KTGain{0.8634}{3.36} & \KTGain{0.7368}{25.54}
& \KTGain{0.7856}{14.54} & \KTGain{0.6482}{10.60} & \KTGain{0.6586}{8.63} & \KTGain{0.6810}{8.44} & \KTGain{0.7107}{7.12} \\
\bottomrule
\end{tabular}\par
\endgroup
\end{table*}

\subsection{Main Results}
\label{sec:main_results}

\noindent\textbf{Performance across selection methods and budgets.}

Table~\ref{tab:main_dbcnn} evaluates GRAM across nine datasets, four selection strategies, and five annotation budgets using DBCNN.
GRAM improves both mean SRCC and PLCC in all 20 strategy--budget configurations.
The gains are particularly pronounced under limited annotation budgets.
For K-Center, the relative improvement in mean SRCC reaches 8.44\% at a 5\% budget, compared with 1.45\% at 50\%.
This highlights the value of quality-aware representations when each selected image must represent a large portion of the candidate pool.
Moreover, GRAM consistently benefits different selection strategies, including K-Center, CD, Moderate, and PGCS, indicating that GRAM improves sample representativeness without requiring changes to the underlying selection algorithms.

% We first test whether quality-aware sample relationships improve subset utility at a fixed budget. Table~\ref{tab:main_dbcnn} pairs Base and GRAM using DBCNN. Both mean SRCC and PLCC improve in all 20 strategy--budget configurations, supporting the utility of the representation across different selection criteria. For K-Center, the relative gain in mean SRCC decreases from 8.44\% at 5\% to 1.45\% at 50\%. Quality-aware distances appear particularly valuable when each selected image must represent a large portion of the pool; broader sampling at larger budgets can partly compensate for a less informative selection space. The pattern depends on the selector: Moderate retains a 3.00\% gain at 50\%, while PGCS gains are smaller overall. This interaction suggests that the representation changes which samples each criterion favors, as examined in Section~\ref{sec:subset_analysis}.

\noindent\textbf{Transfer across downstream models and data domains.}
Table~\ref{tab:kcenter_three_models} fixes K-Center and the 5\% budget and varies the downstream model. GRAM improves mean SRCC by 13.59\%, 6.55\%, and 8.44\% for MANIQA, LIQE, and DBCNN, respectively, with corresponding PLCC gains of 11.65\%, 7.94\%, and 7.12\%. Since GRAM is used only for selection, these gains indicate that the selected samples benefit different IQA architectures. All three models improve in both metrics on CLIVE and AGIQA-3K, suggesting that relationships learned from synthetic distortions also inform selection for authentic and generated images.

\subsection{Analysis of GRM Parameters}
\label{sec:grm-analysis}

% ==================== FIGURE 3 - single column, independent ====================
\begin{figure}[t]
\centering
\captionsetup{type=figure,font=small,skip=3pt,width=\linewidth,justification=justified,singlelinecheck=false}
\includegraphics[width=\linewidth]{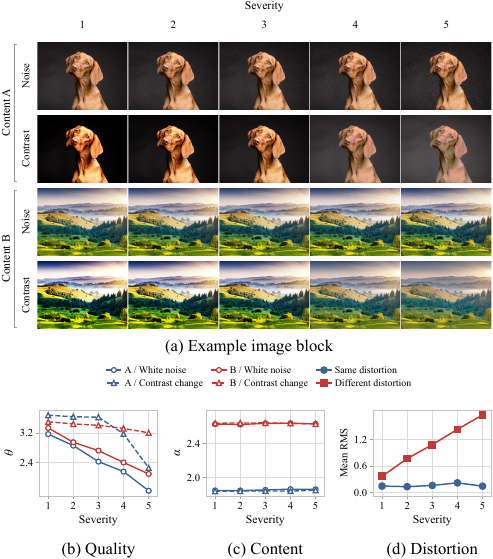}
\caption{Visualization of GRM parameters on KADID. (a) The sampled image block. (b) Predicted $\theta$. (c) Predicted $\alpha$. In both panels, colors indicate image content and line styles indicate distortion types. (d) Mean RMS distances between $\boldsymbol{\beta}$ vectors at matched severity levels: blue compares different contents with the same distortion, and red compares different distortions of the same content. Each point averages two image-pair distances.}
\label{fig:grm-case}
\end{figure}

We analyze 2,000 KADID validation images with reference contents disjoint from upstream training data. Across~400 fixed-content, fixed-distortion sequences, $\theta$ achieves 87.78\% pairwise severity-ranking accuracy. Although ranking supervises the combined response $q(x)$, $\theta$ itself therefore retains a useful quality order. We further construct 36,000 blocks containing two contents and two distortion types. At matched severity, $\alpha$ is closer across distortions of the same content than across contents with the same distortion in 87.46\% of comparisons. For $\boldsymbol{\beta}$, the reverse relation holds in 73.28\% of comparisons under RMS distance. Controlling severity supports the interpretation that these parameters retain complementary content- and distortion-related information.
Fig.~\ref{fig:grm-case} illustrates the same behavior: $\theta$ decreases with severity, $\alpha$ separates contents, and $\boldsymbol{\beta}$ is closer for images sharing a distortion type. These relationships provide multiple cues for distinguishing candidates beyond general visual similarity, supporting the use of the full parameter configuration for selection.

% ==================== FIGURE 5 - gMAD, double column, independent ====================
% \begin{figure*}[!t]
% \begingroup\captionsetup{type=figure,font=small,skip=4pt,width=\textwidth,justification=justified,singlelinecheck=false}
%     \centering
%     \includegraphics[width=0.8\textwidth]{Figure/gMAD.png}
%     \caption{gMAD comparison on KonIQ-10K. DINOv2 and GRAM denote DBCNN models trained on subsets selected using the corresponding representations. (a,b) DINOv2 is fixed at low and high predicted quality levels, respectively. (c,d) GRAM is fixed at low and high predicted quality levels, respectively.}
%     \label{fig:gmad_koniq}
    
% \par\endgroup
% \end{figure*}

\subsection{Subset Distribution and Coverage Analysis}
\label{sec:subset_analysis}

% ==================== FIGURE 4 - single column, independent ====================
\begin{figure}[htbp]
\centering
\captionsetup{type=figure,font=small,skip=3pt,width=\linewidth,justification=justified,singlelinecheck=false}
\includegraphics[width=\linewidth]{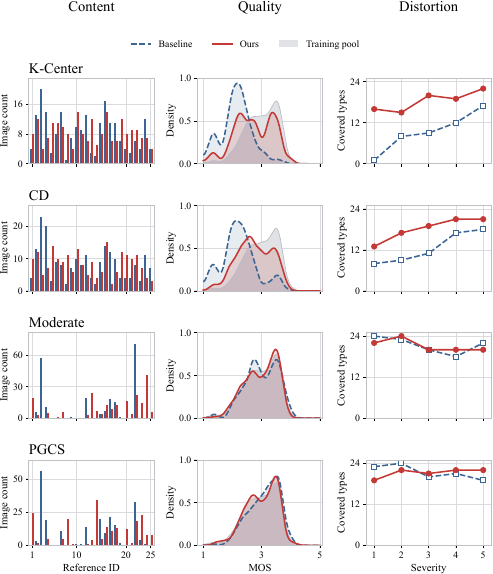}
\caption{Subset distributions and coverage on TID2013 at a 10\% selection budget. Blue: Baseline (DINOv2); red: Ours (GRAM). Left: selected-image counts by reference content. Middle: MOS densities of the selected subsets, with the full training pool in gray. Right: number of covered distortion types at each severity level.}
\label{fig:subset_analysis}
\end{figure}

\begin{figure}[t]
    \centering
    \includegraphics[width=0.5\textwidth]{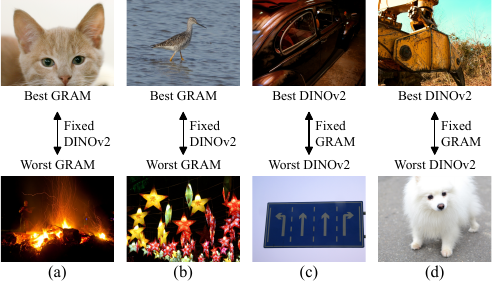}
    \caption{gMAD comparison on KonIQ-10K. DINOv2 and GRAM denote DBCNN models trained on subsets selected using the corresponding representations.}
    \label{fig:gmad_koniq}
    
\end{figure}

Fig.~\ref{fig:subset_analysis} compares the content allocation, quality distribution, and distortion coverage of subsets selected using the two representations. For K-Center and CD, GRAM reduces the concentration of selected images in the low-quality region and covers more distortion types at each severity level. For Moderate and PGCS, the changes are more evident in the more balanced allocation of images across reference contents. Across datasets, selection budgets, and selection strategies, the improvements from GRM augmentation vary across the content, quality, and distortion dimensions.

We further conduct a gMAD comparison~\cite{ma2016gmad} on KonIQ-10K. The two DBCNN models are trained on 10\% subsets selected by K-Center using DINOv2 and GRAM representations, respectively. In Fig.~\ref{fig:gmad_koniq}(a) and \ref{fig:gmad_koniq}(b), when the DINOv2-based model acts as the defender, the GRAM-based model identifies image pairs with visible quality differences. After exchanging their roles, the pairs identified by the DINOv2-based model exhibit less apparent quality differences despite similar scores predicted by the GRAM-based defender, as shown in Fig.~\ref{fig:gmad_koniq}(c) and \ref{fig:gmad_koniq}(d). Visual inspection of these examples suggests that the GRAM-based model performs better in both distinguishing quality differences and resisting attacks, providing qualitative support for its selection effectiveness.

\subsection{Ablation}
\label{sec:ablation}

% ==================== TABLE 3 - single column, independent ====================
\begin{table}[t]
\begingroup\captionsetup{type=table}
\centering
\captionsetup{font=small,skip=3pt,width=\linewidth,justification=justified,singlelinecheck=false}
\caption{GRM parameter ablation with DBCNN, CD selection, and a 10\% training budget. Checkmarks indicate components included in the selection representation. Ours combines DINOv2 with all three GRM components ($\theta$, $\alpha$, and $\beta$). The best result in each column is bold.}
\label{tab:grm_ablation}
\begingroup
\definecolor{AblHeaderBase}{cmyk}{0.62,0.57,0.23,0}
\colorlet{AblHeader}{AblHeaderBase!20}
\colorlet{AblStripe}{gray!10}
\fontsize{6.5}{8.5}\selectfont
\renewcommand{\arraystretch}{1.05}
\setlength{\arrayrulewidth}{0.25pt}
\setlength{\tabcolsep}{1.4pt}
\def\AblMetricWidth{\dimexpr(\linewidth-61pt-20\tabcolsep-\arrayrulewidth)/6\relax}
\begin{tabular}{W{c}{28pt}*{3}{W{c}{11pt}}|*{6}{W{c}{\AblMetricWidth}}}
\Xhline{0.7pt}
\rowcolor{AblHeader}
 &  &  &  & \multicolumn{2}{c}{} & \multicolumn{2}{c}{} & \multicolumn{2}{c}{\textbf{AIGCIQA}} \\
\rowcolor{AblHeader}
 &  &  &  & \multicolumn{2}{c}{\multirow{-2}{*}{\textbf{CSIQ}}} & \multicolumn{2}{c}{\multirow{-2}{*}{\textbf{CLIVE}}} & \multicolumn{2}{c}{\textbf{2023}} \\
\rowcolor{AblHeader}
\multirow{-3}{*}{\textbf{DINOv2}} & \multirow{-3}{*}{$\boldsymbol{\alpha}$} & \multirow{-3}{*}{$\boldsymbol{\beta}$} & \multicolumn{1}{c|}{\multirow{-3}{*}{$\boldsymbol{\theta}$}} & \textbf{SRCC} & \textbf{PLCC} & \textbf{SRCC} & \textbf{PLCC} & \textbf{SRCC} & \textbf{PLCC} \\
\Xhline{0.7pt}
$\checkmark$ & -- & -- & -- & $0.8304$ & $0.8841$ & $0.6327$ & $0.6752$ & $0.7428$ & $0.7554$ \\
\rowcolor{AblStripe}
-- & $\checkmark$ & $\checkmark$ & $\checkmark$ & $0.8711$ & $0.9005$ & $0.6426$ & $0.6811$ & $0.7477$ & $0.7621$ \\
$\checkmark$ & $\checkmark$ & -- & -- & $0.8642$ & $0.8978$ & $0.6490$ & $0.6885$ & $0.7471$ & $0.7618$ \\
\rowcolor{AblStripe}
$\checkmark$ & -- & -- & $\checkmark$ & $0.8598$ & $0.8975$ & $0.6488$ & $0.6896$ & $0.7487$ & $0.7641$ \\
$\checkmark$ & -- & $\checkmark$ & -- & $0.8547$ & $0.8969$ & $0.6494$ & $0.6904$ & $0.7474$ & $0.7632$ \\
\rowcolor{AblStripe}
$\checkmark$ & $\checkmark$ & -- & $\checkmark$ & $0.8567$ & $0.8983$ & $0.6465$ & $0.6897$ & $0.7450$ & $0.7622$ \\
$\checkmark$ & $\checkmark$ & $\checkmark$ & -- & $0.8477$ & $0.8962$ & $0.6501$ & $0.6924$ & $0.7448$ & $0.7614$ \\
\rowcolor{AblStripe}
$\checkmark$ & -- & $\checkmark$ & $\checkmark$ & $0.8407$ & $0.8920$ & $0.6501$ & $0.6917$ & $0.7428$ & $0.7587$ \\
\hline
\multicolumn{4}{c|}{\textbf{Ours}} & $\mathbf{0.8813}$ & $\mathbf{0.9059}$ & $\mathbf{0.6670}$ & $\mathbf{0.7143}$ & $\mathbf{0.7539}$ & $\mathbf{0.7697}$ \\
\Xhline{0.7pt}
\end{tabular}\par
\endgroup
\endgroup
\end{table}

% ==================== TABLE 4 - single column, independent ====================
\begin{table}[t]
\begingroup\captionsetup{type=table}
\centering
\captionsetup{font=small,skip=3pt,width=\linewidth,justification=justified,singlelinecheck=false}
\caption{Effects of combining GRM with different visual representations, using DBCNN, CD selection, and a 10\% training budget. Bold indicates the better result within each pair.}
\label{tab:backbone_ablation}
\begingroup
\definecolor{AblHeaderBase}{cmyk}{0.62,0.57,0.23,0}
\colorlet{AblHeader}{AblHeaderBase!20}
\colorlet{AblStripe}{gray!10}
\fontsize{6.5}{8.5}\selectfont
\renewcommand{\arraystretch}{1.05}
\setlength{\arrayrulewidth}{0.25pt}
\setlength{\tabcolsep}{0.8pt}
\def\AblMetricWidth{\dimexpr(\linewidth-74pt-14\tabcolsep-\arrayrulewidth)/6\relax}
\begin{tabular}{W{l}{74pt}|*{6}{W{c}{\AblMetricWidth}}}
\Xhline{0.7pt}
\rowcolor{AblHeader}
 & \multicolumn{2}{c}{} & \multicolumn{2}{c}{} & \multicolumn{2}{c}{\textbf{AIGCIQA}} \\
\rowcolor{AblHeader}
 & \multicolumn{2}{c}{\multirow{-2}{*}{\textbf{CSIQ}}} & \multicolumn{2}{c}{\multirow{-2}{*}{\textbf{CLIVE}}} & \multicolumn{2}{c}{\textbf{2023}} \\
\rowcolor{AblHeader}
\multicolumn{1}{l|}{\multirow{-3}{*}{\shortstack[l]{\textbf{Selection}\\\textbf{representation}}}} & \textbf{SRCC} & \textbf{PLCC} & \textbf{SRCC} & \textbf{PLCC} & \textbf{SRCC} & \textbf{PLCC} \\
\Xhline{0.7pt}

VGG16 & $0.8138$ & $0.8802$ & $0.6385$ & $0.6853$ & $0.7480$ & $0.7612$ \\
\rowcolor{AblStripe}
VGG16 + GRM & $\mathbf{0.8669}$ & $\mathbf{0.9051}$ & $\mathbf{0.6611}$ & $\mathbf{0.6890}$ & $\mathbf{0.7534}$ & $\mathbf{0.7738}$ \\
\hline

ResNet50 & $0.7730$ & $0.8704$ & $0.6454$ & $0.6899$ & $0.7486$ & $0.7622$ \\
\rowcolor{AblStripe}
ResNet50 + GRM & $\mathbf{0.8711}$ & $\mathbf{0.9056}$ & $\mathbf{0.6671}$ & $\mathbf{0.6942}$ & $\mathbf{0.7519}$ & $\mathbf{0.7671}$ \\
\hline

% SCNN results: to be filled after experiments finish.
SCNN & $0.8308$ & $0.8832$ & $0.6418$ & $0.6742$ & $0.7196$ & $0.7291$ \\
\rowcolor{AblStripe}
SCNN + GRM & $\mathbf{0.8459}$ & $\mathbf{0.8949}$ & $\mathbf{0.6470}$ & $\mathbf{0.6897}$ & $\mathbf{0.7449}$ & $\mathbf{0.7622}$ \\
\hline

LIQE & $0.8561$ & $0.8982$ & $0.6386$ & $0.6967$ & $0.7419$ & $0.7539$ \\
\rowcolor{AblStripe}
LIQE + GRM & $\mathbf{0.8719}$ & $\mathbf{0.9072}$ & $\mathbf{0.6577}$ & $\mathbf{0.6972}$ & $\mathbf{0.7531}$ & $\mathbf{0.7648}$ \\
\hline

DINOv2 (Baseline) & $0.8304$ & $0.8841$ & $0.6327$ & $0.6752$ & $0.7428$ & $0.7554$ \\
\rowcolor{AblStripe}
DINOv2 + GRM (Ours) & $\mathbf{0.8813}$ & $\mathbf{0.9059}$ & $\mathbf{0.6670}$ & $\mathbf{0.7143}$ & $\mathbf{0.7539}$ & $\mathbf{0.7697}$ \\
\Xhline{0.7pt}
\end{tabular}\par
\endgroup
\endgroup
\end{table}

\begin{figure}[t]
    \centering
    \includegraphics[width=\linewidth]{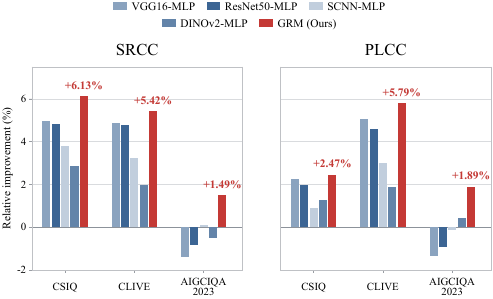}
    \caption{Comparison of GRM with MLP representations trained using only ranking supervision on the same synthetically distorted data. Each representation is combined with DINOv2 features for CD selection. Bars show the percentage changes in SRCC and PLCC relative to the DINOv2 baseline when training DBCNN with a 10\% annotation budget.}
    \label{fig:rank_mlp_ablation}
\end{figure}

We conduct ablation studies on three datasets to examine the contributions of individual GRM parameters and the effects of combining GRM with different visual representations.
As shown in Table~\ref{tab:grm_ablation}, the complete representation achieves the highest SRCC and PLCC on all three datasets and outperforms the configuration using only DINOv2 features. Even when used on their own for selection, the GRM parameters yield
higher SRCC and PLCC than the DINOv2 baseline on all three datasets. Removing any parameter reduces both metrics, supporting the contribution of each parameter to the joint representation.  Table~\ref{tab:backbone_ablation} further shows that adding GRM improves both SRCC and PLCC on all three datasets for DINOv2, ResNet50, SCNN, VGG16, and LIQE representations. These results indicate that the benefits of GRM extend beyond DINOv2 to other visual representations.

To examine whether the gains arise solely from additional training on synthetically distorted data, we train ranking MLPs on different visual features and combine the resulting representations with DINOv2 features for selection. These alternative representations provide some improvements, while GRM achieves higher SRCC and PLCC on all three datasets (Fig.~\ref{fig:rank_mlp_ablation}). This suggests that additional ranking training and feature fusion alone do not fully explain the gains, supporting the added value of GRM modeling.

\section{Conclusion}
We presented GRAM, a psychometrics-inspired framework for IQA coreset selection. Drawing on the graded response model, GRAM represents latent quality together with content-dependent sensitivity and distortion-dependent thresholds, learned from synthetic images without subjective quality supervision. Combining these parameters with visual features improves annotation efficiency across nine datasets, multiple selection strategies, and downstream IQA models. Our analyses support their complementary roles in capturing quality differences. More broadly, our work opens exciting new avenues for interdisciplinary research bridging psychometrics and perceptual quality assessment.

\bibliographystyle{ieeenat_fullname}
\bibliography{main}

@String(ICLR  = {ICLR})

@String(ECCV  = {ECCV})

@String(ICML  = {ICML})

@String(CVPR  = {CVPR})

@String(ICCV  = {ICCV})

@String(CVPRW = {CVPRW})

@String(ICCVW = {ICCVW})

@String(ICIP  = {ICIP})

@String(WACV  = {WACV})

@String(QoMEX = {QoMEX})

@String(CICAI = {CICAI})

@String(TMLR  = {TMLR})

@String(TIP   = {IEEE TIP})

@String(CSVT  = {IEEE TCSVT})

@String(JEI   = {J. Electron. Imaging})

@String(SPIC  = {Signal Process. Image Commun.})

@String(JEBS  = {J. Educ. Behav. Stat.})

@inproceedings{sener2018coreset,
  author    = {Sener, Ozan and Savarese, Silvio},
  title     = {Active Learning for Convolutional Neural Networks: A Core-Set Approach},
  booktitle = ICLR,
  year      = {2018},
  url       = {https://arxiv.org/abs/1708.00489}
}

@inproceedings{xia2023moderate,
  author    = {Xia, Xiaobo and Liu, Jiale and Yu, Jun and Shen, Xu and Han, Bo and Liu, Tongliang},
  title     = {Moderate Coreset: A Universal Method of Data Selection for Real-world Data-efficient Deep Learning},
  booktitle = ICLR,
  year      = {2023},
  url       = {https://openreview.net/forum?id=7D5EECbOaf9}
}

@inproceedings{agarwal2020contextual,
  author    = {Agarwal, Sharat and Arora, Himanshu and Anand, Saket and Arora, Chetan},
  title     = {Contextual Diversity for Active Learning},
  booktitle = ECCV,
  pages     = {137--153},
  year      = {2020},
  url       = {https://arxiv.org/abs/2008.05723}
}

@inproceedings{killamsetty2021gradmatch,
  author    = {Killamsetty, Krishnateja and S, Durga and Ramakrishnan, Ganesh and De, Abir and Iyer, Rishabh},
  title     = {{GRAD-MATCH}: Gradient Matching based Data Subset Selection for Efficient Deep Model Training},
  booktitle = ICML,
  series    = {Proceedings of Machine Learning Research},
  volume    = {139},
  pages     = {5464--5474},
  publisher = {PMLR},
  year      = {2021},
  url       = {https://proceedings.mlr.press/v139/killamsetty21a.html}
}

@inproceedings{cui2026fast,
  author    = {Cui, Jin and Zhao, Boran and Xu, Jiajun and Guo, Jiaqi and Guan, Shuo and Ren, Pengju},
  title     = {{FAST}: Topology-Aware Frequency-Domain Distribution Matching for Coreset Selection},
  booktitle = CVPR,
  pages     = {24748--24758},
  year      = {2026},
  url       = {https://arxiv.org/abs/2511.19476}
}

@inproceedings{nema2025holistic,
  author    = {Nema, Arpita and Zhu, Hanwei and Lin, Weisi},
  title     = {Holistic Coreset Selection for Data Efficient Image Quality Assessment},
  booktitle = ICIP,
  pages     = {2432--2437},
  year      = {2025},
  doi       = {10.1109/ICIP55913.2025.11084586}
}

@inproceedings{nema2026diversity,
  author    = {Nema, Arpita and Zhu, Hanwei and Zhang, Xi and Lin, Weisi},
  title     = {Diversity Preserving Coresets for Image Quality Assessment},
  booktitle = WACV,
  pages     = {7925--7934},
  year      = {2026},
  url       = {https://openaccess.thecvf.com/content/WACV2026/html/Nema_Diversity_Preserving_Coresets_for_Image_Quality_Assessment_WACV_2026_paper.html}
}

@inproceedings{shinde2026wcpc,
  author    = {Shinde, Tushar and Roy, Sreejita},
  title     = {When Every Label Counts: Wasserstein-Optimal Dataset Representation for Coreset Selection in Annotation-Scarce {AI}-Generated Image Quality Assessment},
  booktitle = CVPRW,
  pages     = {2974--2982},
  year      = {2026},
  url       = {https://openaccess.thecvf.com/content/CVPR2026W/DataCV/papers/Shinde_When_Every_Label_Counts_Wasserstein-Optimal_Dataset_Representation_for_Coreset_Selection_CVPRW_2026_paper.pdf}
}

@article{oquab2024dinov2,
  author  = {Oquab, Maxime and Darcet, Timoth{\'e}e and Moutakanni, Th{\'e}o and Vo, Huy V. and Szafraniec, Marc and Khalidov, Vasil and Fernandez, Pierre and Haziza, Daniel and Massa, Francisco and El-Nouby, Alaaeldin and Assran, Mahmoud and Ballas, Nicolas and Galuba, Wojciech and Howes, Russell and Huang, Po-Yao and Li, Shang-Wen and Misra, Ishan and Rabbat, Michael and Sharma, Vasu and Synnaeve, Gabriel and Xu, Hu and J{\'e}gou, Herv{\'e} and Mairal, Julien and Labatut, Patrick and Joulin, Armand and Bojanowski, Piotr},
  title   = {{DINOv2}: Learning Robust Visual Features without Supervision},
  journal = TMLR,
  year    = {2024},
  url     = {https://arxiv.org/abs/2304.07193}
}

@article{zhang2020dbcnn,
  author  = {Zhang, Weixia and Ma, Kede and Yan, Jia and Deng, Dexiang and Wang, Zhou},
  title   = {Blind Image Quality Assessment Using a Deep Bilinear Convolutional Neural Network},
  journal = CSVT,
  volume  = {30},
  number  = {1},
  pages   = {36--47},
  year    = {2020},
  doi     = {10.1109/TCSVT.2018.2886771}
}

@inproceedings{liu2017rankiqa,
  author    = {Liu, Xialei and van de Weijer, Joost and Bagdanov, Andrew D.},
  title     = {{RankIQA}: Learning From Rankings for No-Reference Image Quality Assessment},
  booktitle = ICCV,
  pages     = {1040--1049},
  year      = {2017},
  url       = {https://openaccess.thecvf.com/content_iccv_2017/html/Liu_RankIQA_Learning_From_ICCV_2017_paper.html}
}

@inproceedings{agnolucci2024arniqa,
  author    = {Agnolucci, Lorenzo and Galteri, Leonardo and Bertini, Marco and Del Bimbo, Alberto},
  title     = {{ARNIQA}: Learning Distortion Manifold for Image Quality Assessment},
  booktitle = WACV,
  pages     = {189--198},
  year      = {2024},
  url       = {https://openaccess.thecvf.com/content/WACV2024/html/Agnolucci_ARNIQA_Learning_Distortion_Manifold_for_Image_Quality_Assessment_WACV_2024_paper.html}
}

@inproceedings{saha2023reiqa,
  author    = {Saha, Avinab and Mishra, Sandeep and Bovik, Alan C.},
  title     = {{Re-IQA}: Unsupervised Learning for Image Quality Assessment in the Wild},
  booktitle = CVPR,
  pages     = {5846--5855},
  year      = {2023},
  url       = {https://arxiv.org/abs/2304.00451}
}

@inproceedings{zhang2023liqe,
  author    = {Zhang, Weixia and Zhai, Guangtao and Wei, Ying and Yang, Xiaokang and Ma, Kede},
  title     = {Blind Image Quality Assessment via Vision-Language Correspondence: A Multitask Learning Perspective},
  booktitle = CVPR,
  pages     = {14071--14081},
  year      = {2023},
  url       = {https://arxiv.org/abs/2303.14968}
}

@inproceedings{yang2022maniqa,
  author    = {Yang, Sidi and Wu, Tianhe and Shi, Shuwei and Lao, Shanshan and Gong, Yuan and Cao, Mingdeng and Wang, Jiahao and Yang, Yujiu},
  title     = {{MANIQA}: Multi-Dimension Attention Network for No-Reference Image Quality Assessment},
  booktitle = CVPRW,
  pages     = {1191--1200},
  year      = {2022},
  doi       = {10.1109/CVPRW56347.2022.00126}
}

@article{samejima1969graded,
  author  = {Samejima, Fumiko},
  title   = {Estimation of Latent Ability Using a Response Pattern of Graded Scores},
  journal = {Psychometrika},
  volume  = {34},
  number  = {S1},
  pages   = {1--97},
  year    = {1969},
  doi     = {10.1007/BF03372160}
}

@article{camilli1994scaling,
  author  = {Camilli, Gregory},
  title   = {Teacher's Corner: Origin of the Scaling Constant d = 1.7 in Item Response Theory},
  journal = JEBS,
  volume  = {19},
  number  = {3},
  pages   = {293--295},
  year    = {1994},
  doi     = {10.3102/10769986019003293}
}

@article{sheikh2006live,
  author  = {Sheikh, Hamid R. and Sabir, Muhammad F. and Bovik, Alan C.},
  title   = {A Statistical Evaluation of Recent Full Reference Image Quality Assessment Algorithms},
  journal = TIP,
  volume  = {15},
  number  = {11},
  pages   = {3440--3451},
  year    = {2006},
  doi     = {10.1109/TIP.2006.881959}
}

@article{larson2010csiq,
  author  = {Larson, Eric C. and Chandler, Damon M.},
  title   = {Most Apparent Distortion: Full-Reference Image Quality Assessment and the Role of Strategy},
  journal = JEI,
  volume  = {19},
  number  = {1},
  pages   = {011006},
  year    = {2010},
  doi     = {10.1117/1.3267105}
}

@article{ponomarenko2015tid2013,
  author  = {Ponomarenko, Nikolay and Jin, Lina and Ieremeiev, Oleg and Lukin, Vladimir and Egiazarian, Karen and Astola, Jaakko and Vozel, Benoit and Chehdi, Kacem and Carli, Marco and Battisti, Federica and Kuo, C.-C. Jay},
  title   = {Image Database {TID2013}: Peculiarities, Results and Perspectives},
  journal = SPIC,
  volume  = {30},
  pages   = {57--77},
  year    = {2015},
  doi     = {10.1016/j.image.2014.10.009}
}

@inproceedings{lin2019kadid,
  author    = {Lin, Hanhe and Hosu, Vlad and Saupe, Dietmar},
  title     = {{KADID-10k}: A Large-Scale Artificially Distorted {IQA} Database},
  booktitle = QoMEX,
  pages     = {1--3},
  year      = {2019},
  doi       = {10.1109/QoMEX.2019.8743252}
}

@article{ghadiyaram2016clive,
  author  = {Ghadiyaram, Deepti and Bovik, Alan C.},
  title   = {Massive Online Crowdsourced Study of Subjective and Objective Picture Quality},
  journal = TIP,
  volume  = {25},
  number  = {1},
  pages   = {372--387},
  year    = {2016},
  doi     = {10.1109/TIP.2015.2500021}
}

@article{hosu2020koniq,
  author  = {Hosu, Vlad and Lin, Hanhe and Sziranyi, Tamas and Saupe, Dietmar},
  title   = {{KonIQ-10k}: An Ecologically Valid Database for Deep Learning of Blind Image Quality Assessment},
  journal = TIP,
  volume  = {29},
  pages   = {4041--4056},
  year    = {2020},
  doi     = {10.1109/TIP.2020.2967829}
}

@article{li2024agiqa,
  author  = {Li, Chunyi and Zhang, Zicheng and Wu, Haoning and Sun, Wei and Min, Xiongkuo and Liu, Xiaohong and Zhai, Guangtao and Lin, Weisi},
  title   = {{AGIQA-3K}: An Open Database for {AI}-Generated Image Quality Assessment},
  journal = CSVT,
  volume  = {34},
  number  = {8},
  pages   = {6833--6846},
  year    = {2024},
  doi     = {10.1109/TCSVT.2023.3319020}
}

@inproceedings{wang2024aigciqa,
  author    = {Wang, Jiarui and Duan, Huiyu and Liu, Jing and Chen, Shi and Min, Xiongkuo and Zhai, Guangtao},
  title     = {{AIGCIQA2023}: A Large-Scale Image Quality Assessment Database for {AI} Generated Images: From the Perspectives of Quality, Authenticity and Correspondence},
  booktitle = CICAI,
  pages     = {46--57},
  year      = {2024},
  doi       = {10.1007/978-981-99-9119-8_5}
}

@inproceedings{yuan2025pkuaigiqa,
  author    = {Yuan, Jiquan and Li, Jihe and Yang, Fanyi and Cao, Xinyan and Che, Jinming and Lin, Jinlong and Cao, Xixin},
  title     = {{PKU-AIGIQA-4K}: A Perceptual Quality Assessment Database for Both Text-to-Image and Image-to-Image {AI}-Generated Images},
  booktitle = ICCVW,
  pages     = {3362--3371},
  year      = {2025},
  doi       = {10.1109/ICCVW69036.2025.00353}
}

@misc{mcinnes2018umap,
  author        = {McInnes, Leland and Healy, John and Melville, James},
  title         = {{UMAP}: Uniform Manifold Approximation and Projection for Dimension Reduction},
  howpublished  = {arXiv preprint arXiv:1802.03426},
  year          = {2018},
  eprint        = {1802.03426},
  archivePrefix = {arXiv},
  primaryClass  = {stat.ML}
}

@inproceedings{ma2016gmad,
  author    = {Ma, Kede and Wu, Qingbo and Wang, Zhou and Duanmu, Zhengfang and Yong, Hongwei and Li, Hongliang and Zhang, Lei},
  title     = {Group {MAD} Competition - A New Methodology to Compare Objective Image Quality Models},
  booktitle = CVPR,
  pages     = {1664--1673},
  year      = {2016},
  doi       = {10.1109/CVPR.2016.184}
}

@inproceedings{he2016resnet,
  author    = {He, Kaiming and Zhang, Xiangyu and Ren, Shaoqing and Sun, Jian},
  title     = {Deep Residual Learning for Image Recognition},
  booktitle = CVPR,
  pages     = {770--778},
  year      = {2016}
}

@inproceedings{simonyan2015vgg,
  author    = {Simonyan, Karen and Zisserman, Andrew},
  title     = {Very Deep Convolutional Networks for Large-Scale Image Recognition},
  booktitle = ICLR,
  year      = {2015}
}

\end{document}